\documentclass[sn-mathphys-num]{sn-jnl}

\usepackage{graphicx}%
\usepackage{multirow}%
\usepackage{amsmath,amssymb,amsfonts}%
\usepackage{amsthm}%
\usepackage{mathrsfs}%
\usepackage[title]{appendix}%
\usepackage{xcolor}%
\usepackage{textcomp}%
\usepackage{manyfoot}%
\usepackage{booktabs}%
\usepackage{algorithm}%
\usepackage{algorithmicx}%
\usepackage{algpseudocode}%
\usepackage{listings}%
\usepackage{subcaption}
\usepackage{tabularx}
\usepackage{tabu}

\usepackage{graphicx}  
\usepackage{booktabs}

\usepackage{microtype}
\usepackage{comment}

\usepackage{epigraph} 

\theoremstyle{thmstyleone}%
\theoremstyle{thmstyletwo}%

\theoremstyle{thmstylethree}%

\definecolor{relevant}{RGB}{59, 209, 77}
\definecolor{irrelevant}{RGB}{255, 80, 80}
\definecolor{bbox}{RGB}{45, 214, 214}

\definecolor{QRA}{RGB}{89,156,180}
\definecolor{caption}{RGB}{156, 113, 156}
\definecolor{emotion}{RGB}{79,132,92}

\usepackage[capitalize]{cleveref}
\crefname{section}{Sec.}{Secs.}
\Crefname{section}{Section}{Sections}
\Crefname{table}{Table}{Tables}
\crefname{table}{Tab.}{Tabs.}

\usepackage{amsmath,amsfonts,bm, bbm}

\def\eqref#1{equation~\ref{#1}}

\def\1{\bm{1}}

\DeclareMathAlphabet{\mathsfit}{\encodingdefault}{\sfdefault}{m}{sl}
\SetMathAlphabet{\mathsfit}{bold}{\encodingdefault}{\sfdefault}{bx}{n}

\begin{document}

\title[Article Title]{StimuVAR: Spatiotemporal Stimuli-aware Video Affective Reasoning with Multimodal Large Language Models}


\author*[1]{\fnm{Yuxiang} \sur{Guo}}\email{yguo87@jhu.edu}\equalcont{\small This work was mostly done when Y. Guo was an intern at HRI-USA.}
\author[2]{\fnm{Faizan} \sur{Siddiqui}}
\author[1]{\fnm{Yang} \sur{Zhao}}
\author*[1]{\fnm{Rama} \sur{Chellappa}}\email{rchella4@jhu.edu}
\author*[2]{\fnm{Shao-Yuan} \sur{Lo}}\email{shao-yuan\_lo@honda-ri.com}

\affil[1]{\orgname{Johns Hopkins University}}
\affil[2]{\orgname{Honda Research Institute USA}}

\vspace{-6mm}
\abstract{

Predicting and reasoning how a video would make a human feel is crucial for developing socially intelligent systems. Although Multimodal Large Language Models (MLLMs) have shown impressive video understanding capabilities, they tend to focus more on the semantic content of videos, often overlooking emotional stimuli. Hence, most existing MLLMs fall short in estimating viewers’ emotional reactions and providing plausible explanations. To address this issue, we propose StimuVAR, a spatiotemporal Stimuli-aware framework for Video Affective Reasoning (VAR) with MLLMs. StimuVAR incorporates a two-level stimuli-aware mechanism: frame-level awareness and token-level awareness. Frame-level awareness involves sampling video frames with events that are most likely to evoke viewers’ emotions. Token-level awareness performs tube selection in the token space to make the MLLM concentrate on emotion-triggered spatiotemporal regions. Furthermore, we create VAR instruction data to perform affective training, steering MLLMs’ reasoning strengths towards emotional focus and thereby enhancing their affective reasoning ability. To thoroughly assess the effectiveness of VAR, we provide a comprehensive evaluation protocol with extensive metrics. StimuVAR is the first MLLM-based method for viewer-centered VAR. Experiments demonstrate its superiority in understanding viewers’ emotional responses to videos and providing coherent and insightful explanations.
Our code is available at \href{https://github.com/EthanG97/StimuVAR}{https://github.com/EthanG97/StimuVAR}.
}

\keywords{Video affective reasoning, emotion recognition, emotional stimuli, multimodal large language models}



\maketitle

\section{Introduction}\label{Intro}

\begin{figure}
    \centering
    \includegraphics[width=.9\linewidth]{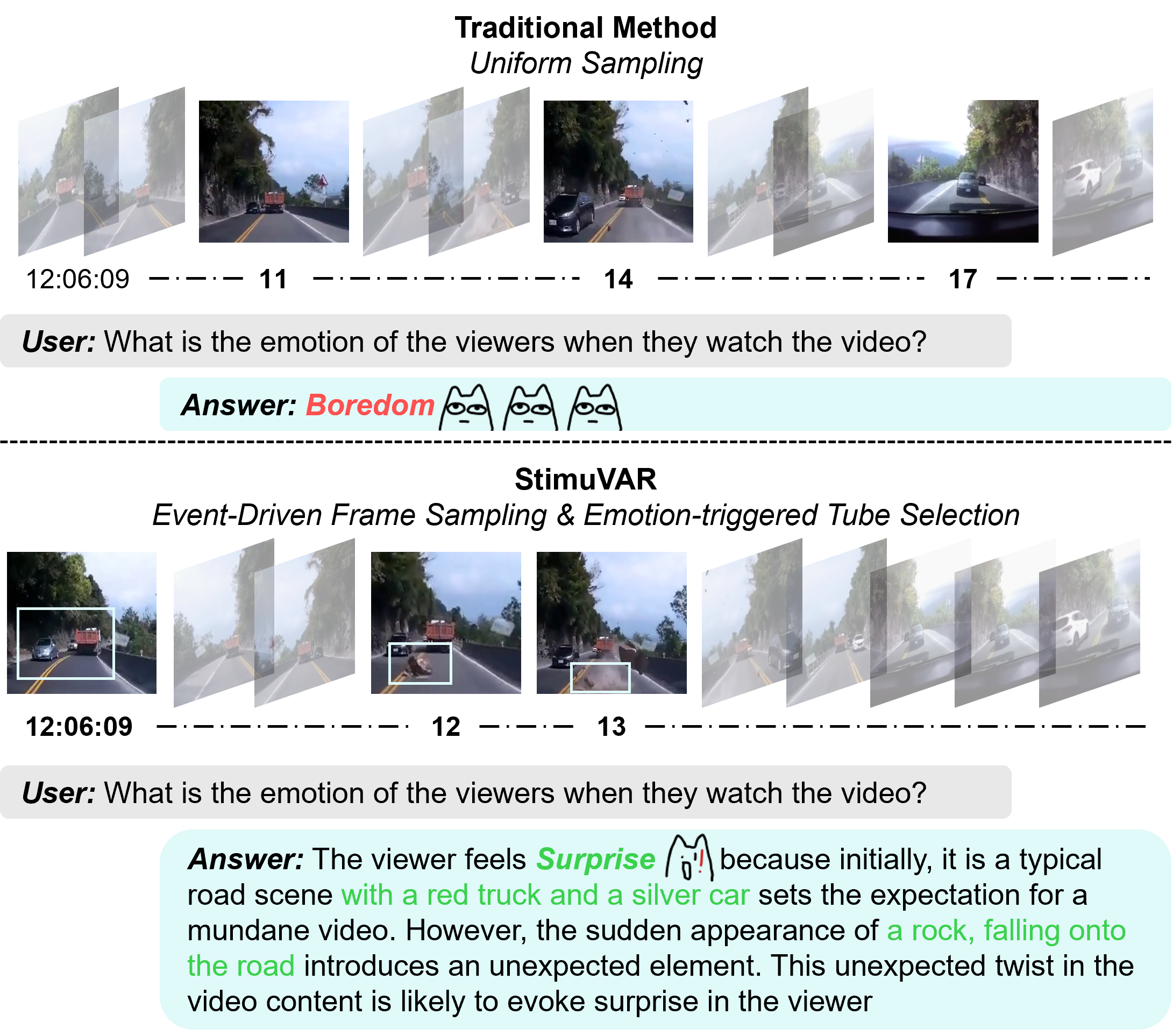}
    \caption{Traditional methods uniformly sample video frames, which could easily miss rapid yet key events that are most likely to evoke viewers' emotions. In contrast, StimuVAR proposes event-driven frame sampling, efficiently selecting the frames containing rapid key events, such as a rock falling onto the road. Next, emotion-triggered tube selection identifies the areas where these events occur, represented by \textcolor{bbox}{color-boxed regions}, guiding MLLM's focus on these emotional stimuli. Additionally, StimuVAR performs affective reasoning, which can offer rationales behind its predictions; for example, it recognizes that the unexpected occurrence of a falling rock triggers the emotion of ``surprise''. \textcolor{relevant}{Relevant} and \textcolor{irrelevant}{irrelevant} words are colored.}
    \label{fig:teaser}
     \vspace{-5mm}
\end{figure}

Understanding human emotional responses to videos is crucial for developing socially intelligent systems that enhance human-computer interaction~\cite{pantic2005affective,wang2020emotion}, personalized services~\cite{bielozorov2019role,lee2017chatbot}, and more. In recent years, user-generated videos on social media platforms have become an integral part of modern society. With increasing concerns about mental health, there is growing public attention on how videos affect viewers' well-being. Unlike most existing Video Emotion Analysis (VEA) approaches that focus on analyzing the emotions of characters in a video~\cite{srivastava2023you,kosti2019context,lee2019context,yang2024robust,lian2023explainable,cheng2024emotion}, predicting and reasoning about a video's emotional impact on viewers is a more challenging task~\cite{achlioptas2021artemis,achlioptas2023affection,mazeika2022would,jiang2014predicting}. This challenge requires not only an understanding of video content but also commonsense knowledge of human reactions and emotions.


Traditional emotion models are trained to map visual embeddings to corresponding emotion labels~\cite{srivastava2023you,kosti2019context,lee2019context,yang2024robust,mazeika2022would,yang2021stimuli,pu2023going}. These models heavily rely on basic visual attributes, such as color, brightness, or object class~\cite{xie2024emovit, yang2023emoset}, which are often insufficient for accurately estimating viewers’ emotional reactions. While recent advances in Multimodal Large Language Models (MLLMs)~\cite{Zhang2023VideoLLaMAAI,Lin2023VideoLLaVALU,luo2023valley,Maaz2023VideoChatGPTTD,li2024mvbench,Jin_2024_CVPR,Ye_2024_CVPR} have demonstrated superiority in various video understanding tasks~\cite{mittal2024can,Wang2023VamosVA,Yang2024FollowTR}, they tend to focus more on the semantic content and factual analysis of videos. This lack of awareness of emotional knowledge often leads these MLLMs to fall short in viewer-centered VEA.


Psychologists have highlighted that emotions are often triggered by specific elements, referred to as emotional stimuli~\cite{mehrabian1974approach,frijda1986emotions,brosch2010perception,peng2016emotions}. \cref{fig:teaser} illustrates an example. Consider a 20-second dashcam video in which a rock suddenly falls onto the road within 2 seconds, while the other 18 seconds depict regular driving scenes. Although most of the video depicts ordinary scenes, the unexpected rock fall is likely to evoke \textit{surprise} and \textit{fear} in viewers. In this case, the falling rock is a critical stimulus that predominantly shapes the viewers' emotional responses. However, current MLLMs may overlook or not prioritize these key emotional stimuli. From a temporal perspective, most MLLMs use uniform temporal downsampling to sample input video frames. While uniform sampling may work well for general video understanding tasks, it could miss the unexpected moments or rapid events that could generate strong reactions and thus make a wrong prediction. From a spatial perspective, emotional stimuli like the falling rock may occupy only a small region of the frame. Identifying these stimulus regions is essential for reducing redundant information and achieving more precise affective understanding.


On the other hand, interpretability is crucial for earning public trust when deploying models in real-world applications. Still, traditional emotion models are not explainable, and most current MLLMs fail to provide plausible affective explanations due to their limited awareness of emotional stimuli, as previously discussed. Although a few recent efforts aim at explainable emotion analysis~\cite{lian2023explainable,cheng2024emotion,achlioptas2021artemis,achlioptas2023affection,xie2024emovit}, they consider only image data or lack a comprehensive evaluation protocol to fully validate their reasoning ability. The task of reasoning human affective responses triggered by videos remains less explored.


To address these limitations, we propose StimuVAR, a spatiotemporal Stimuli-aware framework for Video Affective Reasoning (VAR) with MLLMs. StimuVAR incorporates a two-level stimuli-aware mechanism to identify spatiotemporal stimuli: frame-level awareness and token-level awareness. For frame-level awareness, we introduce \textit{event-driven frame sampling}, using optical flow as a cue to capture the frames that contain unexpected events or unintentional accidents~\cite{epstein2020oops}. These frames are likely to be the stimuli that evoke viewers’ emotions. For token-level awareness, we design \textit{emotion-triggered tube selection} that localizes the emotion-triggered spatiotemporal regions in the token space, which the MLLM can then emphasize. In addition, 
we construct VAR visual instruction data (based on the training set of the large-scale viewer-centered Video Cognitive Empathy (VCE) dataset~\cite{mazeika2022would}) via GPT~\cite{brown2020language,achiam2023gpt} to perform affective training.
The VAR-specific instruction data steer the MLLM's reasoning strengths and commonsense knowledge towards an emotional focus, enhancing the MLLM’s ability to provide insightful and contextually relevant explanations for its affective understanding. To the best of our knowledge, the proposed StimuVAR is the first MLLM-based method for predicting and reasoning viewers’ emotional reactions to videos.


To thoroughly assess the VAR problem, we provide a comprehensive evaluation protocol with extensive metrics, including prediction accuracy, emotional-alignment~\cite{achlioptas2021artemis}, doubly-right~\cite{mao2023doubly}, CLIPScore~\cite{hessel2021clipscore}, and LLM-as-a-judge~\cite{zheng2024judging}. The proposed StimuVAR achieves state-of-the-art performance across multiple benchmarks, demonstrating its ability to predict how a video would make a human feel and to provide plausible explanations. In summary, the main contributions of this work are as follows:

\begin{enumerate}

\item We propose StimuVAR, a novel spatiotemporal stimuli-aware framework for the VAR problem. To the best of our knowledge, it is the first MLLM-based method for predicting and reasoning viewers’ emotional reactions to videos.
\item We propose the event-driven frame sampling and emotion-triggered tube selection strategies to achieve frame- and token-level spatiotemporal stimuli awareness. Furthermore, we create VAR instruction data to perform affective training, enhancing affective reasoning ability.
\item We provide a comprehensive evaluation protocol with extensive metrics for the VAR problem. StimuVAR demonstrates state-of-the-art performance in both prediction and reasoning abilities across multiple benchmarks.
\end{enumerate}

\section{Related Work}\label{Relate}

\subsection{Video Emotion Analysis}
Existing VEA studies can be categorized into two main themes: recognizing the emotions of characters in a video and predicting viewers’ emotional responses to a video. The former primarily relies on facial expressions and dialogue information~\cite{srivastava2023you,kosti2019context,lee2019context,yang2024robust}. We target the latter, which is more challenging as it requires not only video content understanding but also commonsense knowledge of human reactions. Such viewer-centered VEA is crucial for understanding how a video affects human mental health, gaining more attention in the social media era. Traditional supervised approaches learn to map visual features to corresponding emotion labels~\cite{mazeika2022would,yang2021stimuli,pu2023going}. They heavily rely on basic visual attributes, such as color, brightness, or object class~\cite{xie2024emovit}, which are often inadequate for the complex task of viewer-centered VEA. Additionally, these approaches lack interpretability, failing to provide rationales behind their predictions for broader applications. ArtEmis~\cite{achlioptas2021artemis} and Affection~\cite{achlioptas2023affection} introduce benchmarks for viewer-centered affective explanations but only consider image data. Several recent studies explore utilizing MLLMs in emotion analysis. EmoVIT~\cite{xie2024emovit} performs emotion visual instruction tuning~\cite{liu2024visual,Dai2023InstructBLIPTG} to learn emotion-specific knowledge. Although EmoVIT claims to have affective reasoning ability, it does not provide an evaluation. EMER~\cite{lian2023explainable} and Emotion-LLaMA~\cite{cheng2024emotion} use MLLMs to offer explanations for their emotion predictions, but they focus on analyzing characters’ emotions rather than viewers’ emotional responses.

\subsection{Multimodal Large Language Models}
Recent advances in Large Language Models (LLMs), e.g., the GPT~\cite{brown2020language,achiam2023gpt} and LLaMA families~\cite{touvron2023llama1,touvron2023llama,dubey2024llama}, have demonstrated outstanding capabilities in understanding and generating natural languages. These models have been extended to MLLMs~\cite{li2023blip,zhu2024minigpt,liu2024visual,Dai2023InstructBLIPTG} to tackle vision-language problems. For instance, LLaVA~\cite{liu2024visual} and InstructBLIP~\cite{Dai2023InstructBLIPTG} use a projector to integrate visual features into the LLM and conduct visual instruction tuning for instruction-aware vision-language understanding. Several video-oriented MLLMs have also been introduced specifically for video tasks, such as Video-LLaMA~\cite{Zhang2023VideoLLaMAAI}, 
Video-ChatGPT~\cite{Maaz2023VideoChatGPTTD}, 
Chat-UniVi~\cite{Jin_2024_CVPR}, mPLUG-Owl~\cite{Ye_2024_CVPR}, etc. However, these approaches typically employ uniform temporal downsampling to sample input video frames, which often results in missing key stimulus moments that are most likely to trigger viewers’ emotions. In contrast, the proposed StimuVAR incorporates spatiotemporal stimuli awareness, enabling it to achieve state-of-the-art VAR performance.

\section{Method} \label{Method}

\begin{figure}[!t]
    \centering
    \includegraphics[width=.88\linewidth]{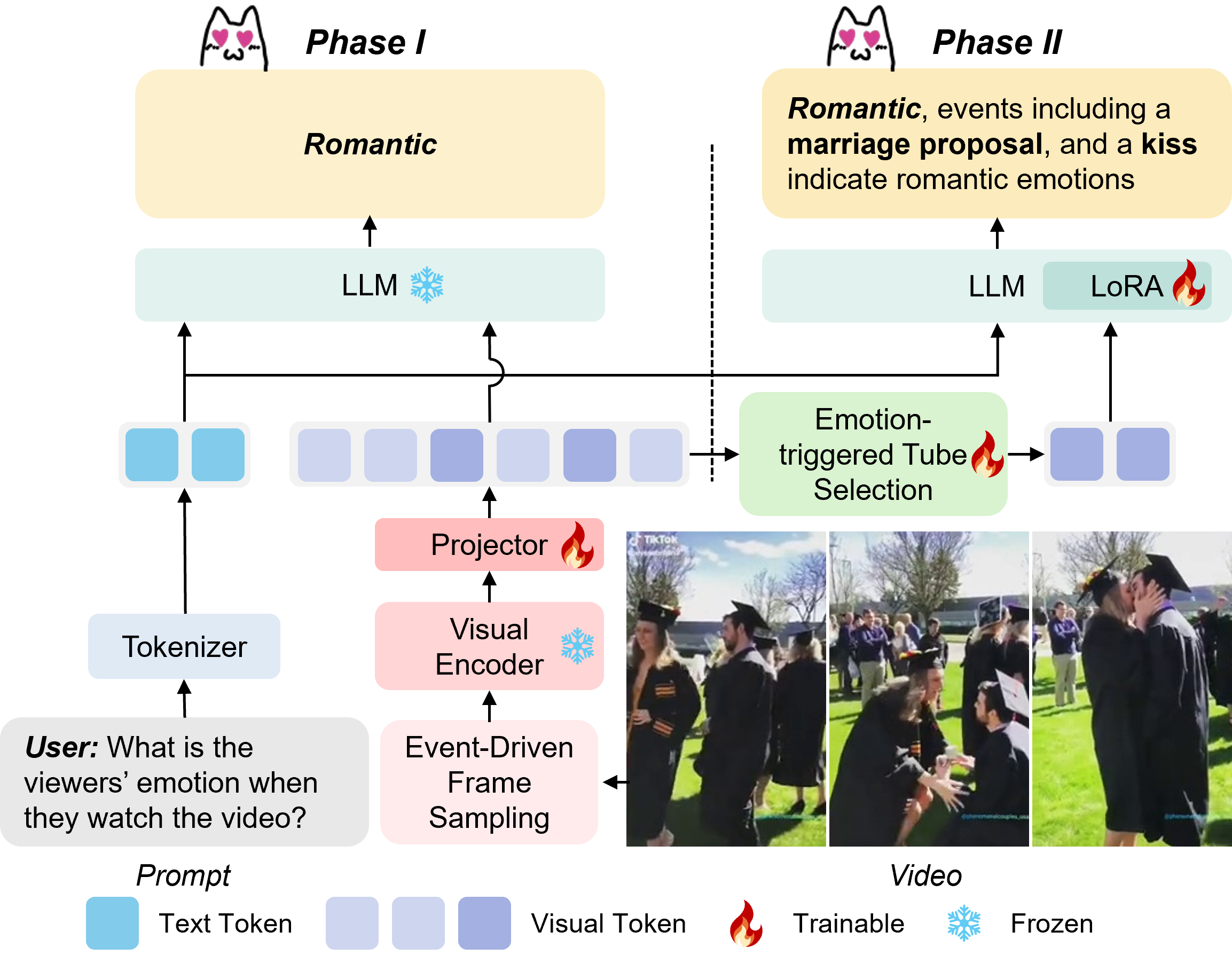}
    \caption{The architecture of StimuVAR. \textit{Event-driven frame sampling} employs optical flow to capture emotional stimuli at the frame level, while \textit{emotion-triggered tube selection} identifies key spatiotemporal areas at the token level, achieving effective and efficient video representations for VAR. To enhance affective understanding, we perform a two-phase \textit{affective training} to steer MLLM's reasoning strengths and commonsense knowledge towards an emotional focus, enabling accurate emotion predictions and plausible explanations.}
    \label{fig:pipeline}
     \vspace{-5mm}
\end{figure}

The proposed StimuVAR, a spatiotemporal stimuli-aware framework for VAR is based on the MLLM backbone. VAR is a task aiming to predict viewers’ emotional responses to a given video and provide reasoning for the prediction. It can be formulated as follows:
\begin{equation} \label{eq:var}
    \{E, R\} = \mathcal{F}(V, P),
\end{equation}
where $V$ is an input video, $P$ is an input text prompt, $E$ is the predicted emotion response, and $R$ is free-form textual reasoning for the emotion prediction $E$. We employ an MLLM as a backbone of the VAR model $\mathcal{F}$. A typical MLLM architecture consists of a visual encoder $\mathcal{F}_{v}$, a projector $\mathcal{F}_{proj}$, a tokenizer $\mathcal{F}_{t}$, and a LLM $\mathcal{F}_{llm}$, and thus Eq.~(\ref{eq:var}) can be written as:
\begin{equation} \label{eq:va2r}
    \{E, R\} = \mathcal{F}_{llm}(\mathcal{F}_{proj}(\mathcal{F}_{v}(V)), \mathcal{F}_{t}(P)).
\end{equation}
The proposed StimuVAR addresses the lack of interpretability in traditional emotion models and the lack of emotional stimuli awareness in existing MLLMs. \cref{fig:pipeline} provides an overview of StimuVAR. The following subsections elaborate on the proposed spatiotemporal stimuli-aware mechanism and affective training.

\subsection{Spatiotemporal Stimuli Awareness} \label{SSA}
The proposed spatiotemporal stimuli-aware mechanism involves two levels of awareness: frame-level and token-level. Frame-level awareness is achieved through the \textit{event-driven frame sampling} strategy, which samples video frames that contain events most likely to evoke viewers’ emotions. Token-level awareness is accomplished via the \textit{emotion-triggered tube selection} strategy, which selects regions in the token space to guide the MLLM’s focus toward emotion-triggered spatiotemporal areas.

\subsubsection{Event-driven Frame Sampling} \label{Event}
In video tasks, uniformly sampling frames is a common practice to represent a video due to temporal redundancy. However, uniform sampling often fails to represent videos containing rapid, unexpected actions or unintentional accidents~\cite{epstein2020oops}, most likely to evoke viewers' emotional reactions~\cite{peng2016emotions}. This is because uniform sampling could easily miss the frames of such rapid but key events. While processing all frames without sampling would preserve all temporal information, the computational burden is significant, especially for MLLMs. To achieve frame-level stimuli awareness, we aim to develop a sampling method that selects the most representative frames within the same constrained number as the uniform sampling baseline. For practical use, we suggest that such frame sampling should meet the following criteria:
\begin{itemize}
    \item \textit{Capture key events}: Identify frames that depict the key events for affective understanding.
    \item \textit{Constrained number of frames}: Select the most representative frames within the allotted frame budget.
    \item \textit{Fast processing}: Perform sampling efficiently to ensure timely analysis.
\end{itemize}
Epstein et al.~\cite{epstein2020oops} study a related task of localizing unintentional actions in videos, exploring cues such as speed and context. Still, their approach requires additional deep networks to extract these video features, making it less ideal for incorporation into the MLLM pipeline when considering processing speed.

\begin{figure}
    \centering
    \includegraphics[width=\linewidth]{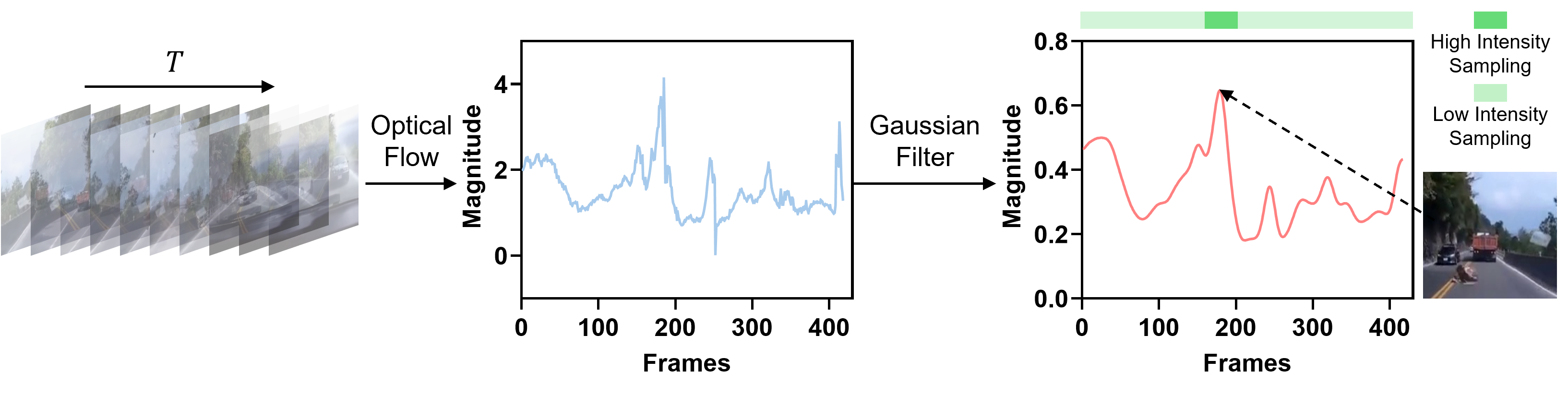}
    \caption{The process of event-driven frame sampling. Optical flows $OF$ between adjacent frames are calculated and then filtered using a Gaussian filter $G_{\sigma}$. Peaks in the filtered optical flows $\tilde{OF}$ are identified as key events $e$. Frames surrounding these events are assigned a high-intensity sampling rate, while the rest of the frames are sampled with a low-intensity rate.}
    \label{fig:frame}
     \vspace{-5mm}
\end{figure}

We propose our event-driven frame sampling strategy based on the observation that rapid key events often coincide with dramatic changes in a video's appearance. These appearance changes can be modeled using the optical flow estimation technique, widely used to capture motions in video tasks~\cite{turaga2009unsupervised, turaga2007videos}.
Let us consider a video $V$ consisting of frames $\{f_1,f_2,...f_T\}$; an optical flow estimator derives the pattern of apparent motion between each pair of adjacent frames $f_t$ and $f_{t+1}$ as follows:
\begin{equation}
    OF_t = \text{OpticalFlow}(f_t, f_{t+1}) \quad \text{for} \quad t = 1, 2, \ldots, T-1, 
\end{equation}
where $OF_t$ is a frame-level optical flow value (i.e., mean absolute of each pixel's optical flow value), and then we obtain a set of estimated optical flows $\{OF_1, OF_2, \ldots, OF_{T-1}\}$ of the video. Although there have been many deep network-based optical flow estimators in the literature~\cite{dosovitskiy2015flownet,ilg2017flownet}, we find that the classic Lucas-Kanade method~\cite{lucas1981iterative} is sufficient for our purpose and offers much higher computational efficiency. As illustrated in \cref{fig:frame}, we construct a curve $OF(t)$ that depicts the intensity of the optical flows over time. To mitigate noise-induced fluctuations, we apply Gaussian smoothing to this curve using a Gaussian filter $G_{\sigma}$:
\begin{equation}
    \tilde{OF}(t) = (OF * G_{\sigma})(t) = \sum_{\tau=-\infty}^{\infty} OF(t-\tau) \, G_{\sigma}(\tau).
\end{equation}
We define the $p$ highest peaks in the smoothed curve $\tilde{OF}(t)$ as the center of key events, i.e., $\{\tilde{OF}_{e1}, \tilde{OF}_{e2}, \ldots, \tilde{OF}_{ep}\}$, where each $e$ denotes a key event in the video. 
The peaks are determined based on a predefined minimum distance between each other and prominence.
Then we locate the corresponding frames $\{f_{e1}, f_{e2}, \ldots, f_{ep}\}$ of the $p$ peaks, and each event is centered around its peak frame, spanning a duration of $2d$ frames as $\{f_{ei-d},\ldots, f_{ei}, \ldots, f_{ei+d}\} \quad \text{for} \quad i = 1, 2, \ldots, p$. We designate the $2d+1$ frames of each event as event-related frames, while the remaining $T - p \times (2d+1)$ non-event frames are collectively treated as a single ``event''.

Given the higher likelihood of important information within the event-related frames, we assign a high-intensity sampling rate to them, while the non-event frames are sampled with a low-intensity rate. Let us consider a predefined number of frames to sample as $N$, these $N$ frames are evenly distributed across all ``events'' $\textbf{e} = \{e_{1}, e_{2}, \ldots, e_{p+1}\}$, where we regard non-event as a single ``event''.
That is, we uniformly sample $\frac{N}{p+1}$ frames from each event set, and more sampled frames are involved in the event set compared to the non-event set, thereby achieving varying sampling rates. This enables discriminative sampling based on event occurrence, ensuring an efficient allocation of resources to effectively capture the essential frames from the video.


\subsubsection{Emotion-triggered Tube Selection} \label{Tube}
After sampling the informative frames that represent a video, we further select the essential regions of interest in the MLLM's token space that are more likely to trigger human emotions, thereby achieving token-level stimuli awareness. This emotion-triggered tube selection module guides the MLLM's focus on the stimulus regions, enhances interpretability, and reduces tokens, leading to a decrease in computational cost.

\begin{figure}
    \centering
    \includegraphics[width=1\linewidth]{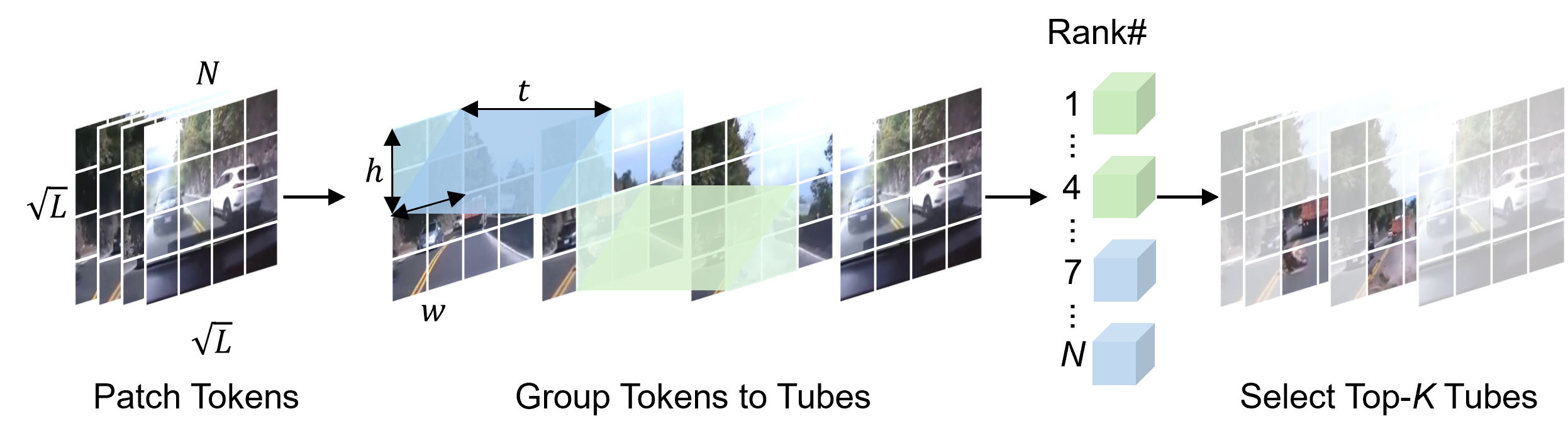}
    \caption{The process of emotion-triggered tube selection. Patch tokens from selected frames are reshaped according to the patch coordinates of the frames and are grouped into tubes of shape $t\times h\times w$. The correlation score estimation determines the importance of each tube for VAR. Then, we select the Top-K tubes with the highest scores as the final video representation.}
    \label{fig:tube}
     \vspace{-5mm}
\end{figure}


Inspired by patch selection in visual recognition~\cite{cordonnier2021differentiable}, we formulate token selection as a Top-K problem.
As illustrated in \cref{fig:pipeline}, to identify emotional stimuli areas in the token space, we focus on the \texttt{<Patch>} tokens of the visual tokens after the projector, denoted as $\textbf{k} \in \mathbb{R}^{N\times L \times C}$, where $C$ is the embedding dimension and $L$ is the number of tokens for each frame. We use a two-layer perceptron $\mathcal{S}$ to estimate correlation scores between each token and the output response, given by $\textbf{c} = \mathcal{S}(\textbf{k}) \in \mathbb{R}^{N\times L}$. Next, we reshape the correlation scores $\textbf{c}$ into a 3D volume $\textbf{c}^{r} \in \mathbb{R}^{N\times \sqrt{L} \times \sqrt{L}}$ according to the patch coordinates of the original input frames. This volume is split into tubes of shape $t\times h \times w$, resulting in $((\frac{N-t}{d_{t}}+1)\times (\frac{\sqrt{L}-h}{d_{h}}+1) \times (\frac{\sqrt{L}-w}{d_{w}}+1))$ tubes, where $(d_{t}, d_{w}, d_{h})$ denotes the stride shape. The token scores within each tube are averaged to obtain each tube's score, and then we select the Top-K tubes with the highest correlation scores as the final video representation sent to the following LLM.

This design differs from existing token selection approaches~\cite{wang2022efficient} that consider the temporal and spatial domains sequentially. In these approaches, temporal selection highly influences the final performance, especially after the frame sampling step. Missing essential frames that trigger emotions during the token selection process significantly reduces the chance of making a correct prediction. We provide empirical evidence in \cref{fig:temporal_module} to support this claim. In contrast, our tube selection strategy groups tokens into tubes, ensuring that the selection is driven by both temporal and spatial information. This design preserves each frame's intrinsic spatial structure and accounts for the entire video's consistency and continuity. Furthermore, our tube selection allows for efficient utilization of computational resources while identifying key emotional stimuli in a video.

\subsection{Affective Training} \label{affective_training}

To fully integrate the proposed stimuli-aware mechanism into the MLLM backbone and to enhance the MLLM's affective reasoning ability, we introduce an affective training protocol to fine-tune the MLLM.

\subsubsection{VAR Visual Instruction Data Construction}
\label{var_data}

To prepare effective affective training, we construct VAR-specific visual instruction data based on the training set of the raw Video Cognitive Empathy (VCE) dataset~\cite{mazeika2022would}, a large-scale viewer-centered video dataset with fine-grained annotations. Detailed dataset information is provided in \cref{dataset}. Given the strong capabilities of GPT~\cite{brown2020language,achiam2023gpt} for constructing logical reasoning across concepts, we leverage it to generate instructions with causal relationships between videos and the emotions they evoke. 

Due to GPT’s limitation in directly processing large-scale video data, we first employ a vision-language model\cite{wang2023cogvlm} to generate captions for each sampled frame, structuring them sequentially in a format of ``Frame 1 description: ...; Frame 2 description: ...''. GPT then processes these frame-level captions, incorporating temporal correlations to produce a coherent video-level caption. Compared to directly captioning videos in a single step using a video-oriented MLLM~\cite{Zhang2023VideoLLaMAAI,Lin2023VideoLLaVALU}, this progressive summarization method captures video details and ensures frame-level temporal consistency, thereby mitigating hallucination.

Next, inspired by~\cite{zhang2024mm}, we prompt GPT to generate a reasoning process of deriving the label from an input. Specifically, given the pairs of the video caption \texttt{<\textcolor{caption}{Video Caption}>} (i.e., the caption of $V$) and the emotional response label \texttt{<\textcolor{emotion}{Emotion}>} (i.e., $Y_E$), we query GPT to explain viewers' emotional responses when they watch the videos.
This approach enables us to generate detailed video captions along with a structured rationale that explicitly connects viewers' emotional responses to the underlying stimuli. Examples of video captions and instructions are provided in Supplementary Material. An example of our prompt is shown below, and its output reasoning is denoted as $Y_R$:

\begin{minipage}{0.95\textwidth}
\begin{lstlisting}
    {``role'': ``system'',
     ``content'': Given the below (QUESTION, ANSWER) pair examples of emotion estimation, left fill-in the REASONING process which derives ANSWERS from QUESTIONS in three sentences.},
    {``role'': ``user'', 
     ``content'': QUESTION: These are frame descriptions from a video. After reading the descriptions, how people might emotionally feel about the content and why. Only provide the one most likely emotion. <Video Caption>
     ANSWER: The viewer feels <Emotion>.
     REASONING: Let's think of step-by-step
\end{lstlisting}
\end{minipage}

\subsubsection{Two-phase Training}

Our affective training consists of two phases (see \cref{fig:pipeline}). In the initial phase, we use the original training set of the VCE dataset with videos and emotional response label pairs $\{V, Y_E\}$, 
where only the projector $\theta_{\mathcal{F}_{proj}}$ of the MLLM is trainable, while keeping the pre-trained visual encoder $\mathcal{F}_{v}$ and LLM $\mathcal{F}_{llm}$ frozen, as follows:
\begin{equation}
    \min_{\theta_{\mathcal{F}_{proj}}} \mathcal{L}(\mathcal{F}(V, P; \theta_{\mathcal{F}_{v}}, \theta_{\mathcal{F}_{proj}}, \theta_{\mathcal{F}_{llm}}), Y_E),
\end{equation}
where $\mathcal{L}$ is the cross-entropy loss. 
During this phase, the projector aligns the visual features extracted by the pre-trained visual encoder with the pre-trained LLM word embeddings and learns the correlation between videos and the emotions they trigger, supervised by the ground-truth emotion annotations $Y_E$.


\color{black}

In the second phase, we enhance the MLLM’s ability to offer plausible reasoning for its emotional predictions through visual instruction tuning~\cite{xie2024emovit,liu2024visual,Dai2023InstructBLIPTG}.
We use our VAR instruction data $\{V, Y_E, Y_R\}$ (based on the original training set of the VCE dataset $\{V, Y_E\}$) constructed in \cref{var_data}
to train the proposed emotion-triggered tube selection module $\theta_{tube}$ and fine-tune the projector $ \theta_{\mathcal{F}_{proj}}$ and the LLM (using LoRA~\cite{hu2022lora}) $\theta^{lora}_{\mathcal{F}_{llm}}$. The training objective is formulated as follows:
\begin{equation}
    \min_{\theta_{tune}} \mathcal{L}(\mathcal{F}(V, P; \theta_{\mathcal{F}_{v}}, \theta_{\mathcal{F}_{proj}}, \theta_{tube}, \theta_{\mathcal{F}_{llm}}), \{Y_E, Y_R\}), \quad \theta_{tune} = \{\theta_{\mathcal{F}_{proj}}, \theta_{tube}, \theta^{lora}_{\mathcal{F}_{llm}}\},
\end{equation}
where $\mathcal{L}$ is the cross-entropy loss. 
This fine-tuning enhances the model’s affective reasoning capabilities for its emotional predictions. Furthermore, the emotion-triggered tube selection module strengthens the link between visual tokens of stimuli and the elements referenced in the reasoning, while also filtering out irrelevant visual tokens, reducing noise, improving decision-making, and enhancing the model's efficiency.

Our two-phase affective training protocol directs the MLLM's reasoning strengths and commonsense knowledge towards an emotional focus, enabling the MLLM to offer plausible explanations for its affective understanding.

\section{Experiments}\label{Exp}

\subsection{Experimental Setup} \label{dataset}

\textbf{Datasets.}
We evaluate our method on three viewer-centered VEA datasets. \textit{Video Cognitive Empathy (VCE)}~\cite{mazeika2022would} contains 61,046 videos with 27 fine-grained emotional responses~\cite{cowen2017self} manually annotated. The dataset is divided into 50,000 videos for training and 11,046 for testing, with each video averaging 14.1 seconds in length. This dataset purposely removes audio cues to focus solely on visual information. \textit{VideoEmotion-8 (VE-8)}~\cite{jiang2014predicting} comprises 1,101 user-generated videos sourced from YouTube and Flickr. Each video is categorized into one of eight emotions from Plutchik’s Wheel of Emotions~\cite{plutchik1980general} and has an average duration of around 107 seconds. \textit{YouTube/Flickr-EkmanSix (YF-6)}~\cite{xu2016heterogeneous} shares the same video source as VE-8 but expands the size to 1,637, categorized into Ekman’s six basic emotions~\cite{ekman1971constants}. Since our VAR instruction data is constructed using the training set of VCE, we employ the VCE test set for in-domain evaluation while using VE-8 and YF-6 as out-of-domain datasets to assess our method's effectiveness and generalization capabilities.

\noindent\textbf{Evaluation metrics.}
We follow the official metrics used by the VCE, VE-8 and YF-6 datasets. VCE utilizes the Top-3 accuracy, while VE-8 and YF-6 evaluate the Top-1 accuracy. Additionally, we provide a comprehensive protocol with extensive metrics to evaluate the reasoning quality for the VAR problem. All the metrics are reference-free since existing viewer-centered VEA datasets~\cite{mazeika2022would,jiang2014predicting} do not contain ground-truth reasoning. These metrics are described as follows:
\begin{enumerate}[(i)]
    \item \textit{Emotional-alignment (Emo-align)}~\cite{achlioptas2021artemis} uses GPT-3.5~\cite{brown2020language} as a text-to-emotion classifier $C_{E|R}$ that measures the accuracy (\%) of the emotion predicted from reasoning text. This metric evaluates if the output reasoning $R$ evokes the same emotion $E$ as the ground truth $Y_E$.
    \item \textit{Doubly-right \{RR, RW, WR, WW\}}~\cite{mao2023doubly} jointly considers the accuracy of prediction and reasoning, where RR denotes the percentage (\%) of Right emotional prediction with Right reasoning, RW denotes Right prediction with Wrong reasoning, WR denotes Wrong prediction with Right reasoning, and WW denotes Wrong prediction with Wrong reasoning. Here the correctness of reasoning is defined by Emo-align. We desire a high accuracy of RR (the best is 100\%) and low percentages of RW, WR and WW (the best is 0\%). The sum of \{RR, RW, WR, WW\} is 100\%.
    \item \textit{CLIPScore (CLIP-S)}~\cite{hessel2021clipscore} uses CLIP~\cite{radford2021learning} to measure the compatibility between input videos and output reasoning in feature space. A higher score indicates better video-reasoning compatibility.

    \item \textit{LLM-as-a-judge}~\cite{zheng2024judging} uses GPT-3.5~\cite{brown2020language} and Claude 3.5~\cite{anthropic2024claude} as a judge respectively to rate the quality of reasoning. We set the score as an integer within a range of $[1, 4]$. In addition to average scores, we also report the 1-vs.-1 (StimuVAR vs. a baseline) comparison results showing the number of test data where StimuVAR wins, loses or ties.
\end{enumerate}

\noindent \textbf{Baseline approaches.}
To our knowledge, StimuVAR may be the first method for the viewer-centered VAR problem. Hence, we broadly compare it with traditional zero-shot approaches~\cite{mazeika2022would,radford2021learning}, traditional emotion models~\cite{tran2018closer,sharir2021image,tong2022videomae,pu2023going}, state-of-the-art video-oriented MLLMs~\cite{Zhang2023VideoLLaMAAI,Lin2023VideoLLaVALU,luo2023valley,Maaz2023VideoChatGPTTD,li2024mvbench,Jin_2024_CVPR,Ye_2024_CVPR}, and a long-video MLLM, LongVLM~\cite{weng2024longvlm}. The traditional emotion models are specifically trained for emotion tasks, but they cannot offer the rationales behind their predictions. We also consider EmoVIT~\cite{xie2024emovit}, a state-of-the-art emotion MLLM for image data, and we extend it to videos for comparison.

\noindent \textbf{Implementation details.}
The MLLM backbone uses CLIP ViT-L/14~\cite{radford2021learning,dosovitskiy2021image} as the visual encoder, Llama2-7b~\cite{touvron2023llama} as the LLM, and a linear layer as the projector. For VAR instruction data creation, we employ CogVLM-17B~\cite{wang2023cogvlm} to generate captions for each selected frame and GPT-4~\cite{achiam2023gpt} to aggregate these frame-level captions into a video caption. GPT-3.5~\cite{brown2020language} is then used to generate causal connections between the videos and the emotions they trigger. In event-driven frame sampling, we set $N=6$ as the default number of sampled frames. The emotion-triggered tube selection module refines the \texttt{<Patch>} tokens after the projector, retaining the top $K=4$ tubes with shape $2\times4\times4$ as spatial tokens and \texttt{<CLS>} as temporal tokens to represent a video. The model is trained using the AdamW optimizer~\cite{loshchilov2019decoupled} with a learning rate of $5\times10^{-5}$ and a batch size of $1$, through $2$ and $3$ epochs in the phase I and II of affective training, respectively.

\renewcommand{\arraystretch}{0.99}
\begin{table}[t!]
    \centering
    \setlength{\tabcolsep}{5.5pt}
    \caption{Quantitative comparison on the VCE dataset.}
    \label{tab:VCE_main}
    \begin{tabular}{ll | r | r | rrrr | r}
    \toprule
       Method & Venue &  Top-3 & Emo-align & RR & RW & WR & WW & CLIP-S\\
    \midrule
       \textbf{Traditional} & & & & & & & &\\
       CLIP~\cite{radford2021learning} & ICML'21 &  28.4 & - & - & - & - & - & -\\
       Majority~\cite{mazeika2022would} & NeurIPS'22 &  35.7 & - & - & - & - & - & -\\
       R(2+1)D~\cite{tran2018closer} & CVPR'18 & 65.6 & - & - & - & - & - & -\\
       STAM~\cite{sharir2021image} & arXiv'21 &   66.4 & - & - & - & - & - & -\\
       VideoMAE~\cite{tong2022videomae} & NeurIPS'22 & 68.9 & - & - & - & - & - & -\\
       MM-VEMA~\cite{pu2023going} & PRCV'23 & 73.3 & - & - & - & - & - & -\\
    \midrule
       \textbf{MLLM} & & & & & & & & \\
       Video-LLaMA~\cite{Zhang2023VideoLLaMAAI} & EMNLP'23 &26.4 & 25.5 & 16.2 & 10.2 & 9.3 & 64.3 & 63.9\\
       Video-LLaVA~\cite{Lin2023VideoLLaVALU} & arXiv'23 &25.0 & 31.2 &17.5 & 7.5 & 13.7 & 61.3 & 70.6\\
       Valley~\cite{luo2023valley} & arXiv'23 & 31.3 & 29.4 & 19.2 & 12.1 & 10.2 & 58.5 & 69.4\\   
       Video-ChatGPT~\cite{Maaz2023VideoChatGPTTD} & ACL'24 & 21.0 & 29.5 & 11.4 & 9.5 & 18.1 & 61.0 & 68.9\\
       VideoChat2~\cite{li2024mvbench} & CVPR'24 & 31.1 & 36.4 & 24.0 & 7.1 & 12.4 & 56.5 & 68.6\\
       Chat-UniVi~\cite{Jin_2024_CVPR} & CVPR'24 &  38.6 &29.5 & 21.0 & 17.6 & 8.5 & 52.9  & 70.2\\
       mPLUG-Owl~\cite{Ye_2024_CVPR} & CVPR'24 & 23.6 & 22.1 & 13.8 & 9.7 & 8.3 & 68.2 & 69.3\\
       EmoVIT~\cite{xie2024emovit} & CVPR'24 & 10.5 & 5.2 & 4.8 & 5.7 & 0.4 & 89.1 & 48.9\\    
       LongVLM~\cite{weng2024longvlm} & ECCV'24 &20.9 & 19.7 &6.9 & 13.9 & 12.8 & 66.4 & 64.3\\
    \midrule    
       StimuVAR (Ours) &  &  \textbf{73.5} & \textbf{69.6} & \textbf{68.8} & 4.7 & 0.8 & 25.6 & \textbf{75.3}\\
    \bottomrule
    \end{tabular}
\end{table}

\begin{table}[t!]
    \centering
    \setlength{\tabcolsep}{9pt}
    \caption{LLM-as-a-judge comparison on the VCE and VE-8 datasets. Due to GPT's query limitation, 100 test videos are randomly selected for evaluation. ``Win/Lose/Tie'' denotes the number of samples that StimuVAR wins/loses/ties against each competitor in a 1-vs.-1 comparison. ``Score'' denotes the average score on the 100 test samples.}
    \begin{tabular}{l | rrrr | rrrr}
    \toprule
        Datasets & \multicolumn{4}{c !{\vrule width 1pt}}{VCE~\cite{mazeika2022would}} & \multicolumn{4}{c}{VE-8~\cite{jiang2014predicting}}\\
        \midrule
        Judgement & Win & Lose & Tie & Score & Win & Lose & Tie & Score\\
        \midrule
        Video-LLaMA~\cite{Zhang2023VideoLLaMAAI} & 41 & 11 & 48 & 3.03 &48 & 13 & 39 & 2.83\\
        Video-LLaVA~\cite{Lin2023VideoLLaVALU} & 45 & 13 & 42 & 3.02 & 49 & 23 & 28 & 2.45\\
        Valley~\cite{luo2023valley} & 45 & 12 & 43 & 2.89 & 35 & 7 & 58 & 2.88\\
        Video-ChatGPT~\cite{Maaz2023VideoChatGPTTD} & 54 & 8 & 38 & 2.69 & 30 & 18 & 52 & 3.12\\
        VideoChat2~\cite{li2024mvbench} & 56 & 12 & 32 & 2.65 & 48 & 13 & 39 & 3.02\\
        Chat-UniVi~\cite{Jin_2024_CVPR} & 50 & 19 & 31 & 2.83 & 37 & 20 & 43 & 3.08\\
        mPLUG-Owl~\cite{Ye_2024_CVPR} & 37 & 17 & 46 & 3.18 & 29 & 20 & 51 & 3.14\\
        EmoVIT~\cite{xie2024emovit} & 79 & 2 & 19 & 2.18 & 92 & 0 & 8 & 1.99\\
        \midrule
        StimuVAR (Ours) & - & - & - & \textbf{3.40} & - & - & - & \textbf{3.29}\\
    \bottomrule
    \end{tabular}
    
    \label{tab:judge_LLM}

\end{table}

\subsection{Main Results}
\Cref{tab:VCE_main} reports the results on the VCE dataset. Traditional emotion models generally outperform MLLMs in prediction accuracy since they are specifically trained for emotion tasks. However, these traditional models are not able to provide explanations for their predictions. In contrast, MLLMs have a significant reasoning capability. Compared to existing video-oriented MLLMs, the proposed StimuVAR demonstrates superiority in the VAR task, both in predicting viewers' emotional responses and offering rationales behind its predictions. We can observe that StimuVAR achieves 69.6\% in Emo-align, which is double the performance of most MLLM baselines. It also has 68.8\% in RR, which shows consistency between emotional predictions and reasoning. StimuVAR's highest CLIP-S score indicates a better correspondence between input videos and output reasoning, suggesting that its reasoning closely adheres to the visual cues and covers details. Notably, while EmoVIT~\cite{xie2024emovit} is an MLLM specifically trained with emotion knowledge, it does not perform well on viewer-centered VEA datasets. The reason may be that EmoVIT prioritizes image attributes, whereas video data require temporal attention. This validates the importance of a spatiotemporal stimuli-aware mechanism. \Cref{tab:judge_LLM} presents the LLM-as-a-judge results using GPT. Due to GPT's query limitation, we randomly select 100 test videos for this metric. StimuVAR achieves the highest GPT rating scores and better winning percentages against all competitors in 1-vs.-1 comparisons, demonstrating its coherent and insightful explanations. The results using Claude 3.5 as a judge are reported in the Supplementary Material.

\Cref{tab:judge_LLM} and \Cref{tab:ve8} present the results on the VE-8 dataset. To test the generalizability of StimuVAR, we evaluate the model trained with VCE data without fine-tuning on VE-8. Although VE-8 has a different emotion categorization, lower video quality, and nearly eight times longer video duration, StimuVAR again outperforms all the MLLM baselines. Furthermore, this performance is achieved by sampling only six frames, making the computational efficiency more significant for these long videos. \Cref{tab:yf6} shows a performance ranking in the YF-6 dataset similar to that of VE-8, where StimuVAR achieves the highest Top-1 accuracy of 46.2\%.
These results demonstrate that the proposed emotional stimuli awareness has high generalizability and can efficiently reduce computational costs. Overall, StimuVAR consistently achieves the best scores across all the metrics on both datasets.

\begin{table}[t!]
    \centering
    \setlength{\tabcolsep}{4.5pt}
    \caption{Quantitative comparison on the VE-8 dataset.}
    \label{tab:ve8}
    \begin{tabular}{ll | r | r | rrrr | r}
    \toprule
       Method & Venue & Accuracy & Emo-align & RR & RW & WR & WW & CLIP-S\\
\midrule
       Video-LLaMA~\cite{Zhang2023VideoLLaMAAI} & EMNLP'23 &12.4 & 21.6 & 10.4 & 1.9 & 11.3 & 76.4 & 62.3\\
       Video-LLaVA~\cite{Lin2023VideoLLaVALU} & arXiv'23 & 21.5 & 20.5 &17.7 & 3.8 & 2.8 & 75.7 & 61.3\\
       Valley~\cite{luo2023valley} & arXiv'23 & 16.1  & 20.6 & 12.2 & 4.0 & 8.5 & 75.3 & 67.0\\ 
       Video-ChatGPT~\cite{Maaz2023VideoChatGPTTD} & ACL'24 & 17.2  & 16.6 & 14.1 & 3.1 & 2.5 & 80.2 & 66.0\\
       VideoChat2~\cite{li2024mvbench} & CVPR'24 & 16.5  & 15.6 & 14.3 & 2.2 & 1.4 & 82.1 & 65.1\\
       Chat-UniVi~\cite{Jin_2024_CVPR} & CVPR'24 & 14.4  &19.2 & 9.9 & 4.5 & 9.3 & 76.3  & 65.9\\
       mPLUG-Owl~\cite{Ye_2024_CVPR} & CVPR'24 & 26.2  & 24.7 & 19.1 & 7.0 & 5.6 & 68.3 & 64.4\\
       EmoVIT~\cite{xie2024emovit} & CVPR'24 & 25.1  & 0.3 & 0.3 & 24.8 & 0.0 & 74.9 & 45.9\\       
    \midrule    
       StimuVAR (Ours) &  &  \textbf{40.5}  & \textbf{41.7} & \textbf{33.7} & 6.8 & 8.0 & 51.5 & \textbf{70.6}\\
    \bottomrule
    \end{tabular}
\end{table}

\begin{table}[htb]
    \centering
    \setlength{\tabcolsep}{3.5pt}
    \caption{Quantitative comparison on the YF-6 dataset.}
    \label{tab:yf6}
    \begin{tabular}{l |r | r |  rrrr | r | rrrr}
    \toprule
       Method & Acc & Emo-ali & RR & RW & WR & WW & CLIP-S & Win & Lose & Tie &Score\\
\midrule
       Video-LLaMA~\cite{Zhang2023VideoLLaMAAI}
       & 17.3 & 17.7 & 10.5 & 6.8 & 7.2 & 75.5 & 62.4&44 & 12 & 44 & 2.80\\
       Video-LLaVA~\cite{Lin2023VideoLLaVALU} 
       &  20.1 & 17.5 &17.1 & 3.0 & 0.4 & 79.5 & 66.8& 42 & 17 & 41 & 2.85\\
       Valley~\cite{luo2023valley} &  10.0  & 18.9 & 5.9 & 4.1 & 13.0 & 77.0 & 66.4& 44 & 10 & 46 & 2.89\\ 
       Video-ChatGPT~\cite{Maaz2023VideoChatGPTTD} 
       &  17.9  & 19.7 & 13.2 & 4.7 & 6.5 & 75.6 & 63.8& 43 & 16 & 41 & 2.93\\
       VideoChat2~\cite{li2024mvbench} 
       & 24.3  & 23.7& 18.5 & 5.8 & 5.2 & 70.5 & 64.1& 37 & 16 & 47 & 3.03\\
       Chat-UniVi~\cite{Jin_2024_CVPR} 
       &  28.4  &29.2 & 22.1 & 6.3 & 7.1 & 64.5  & 66.3& 33 & 14 & 53 & 3.12\\
       mPLUG-Owl~\cite{Ye_2024_CVPR} 
       &  22.8  & 24.9 & 14.9 & 7.9 & 10.0 & 67.2 & 53.7& 31 & 17 & 52 & 3.11\\
       EmoVIT~\cite{xie2024emovit} 
       & 0.7  & 1.7 & 0.5 & 0.2 & 1.2 & 98.1 & 52.5& 85 & 0 & 15 & 1.88\\       
    \midrule    
       StimuVAR (Ours) & \textbf{46.2}  & \textbf{44.2} & \textbf{37.3} & 8.9 & 6.9 & 46.9 & \textbf{72.1}& - & - & - & \textbf{3.30}\\
    \bottomrule
    \end{tabular}
\end{table}

\begin{figure}[!t]
    \centering
    \includegraphics[width=1\linewidth]{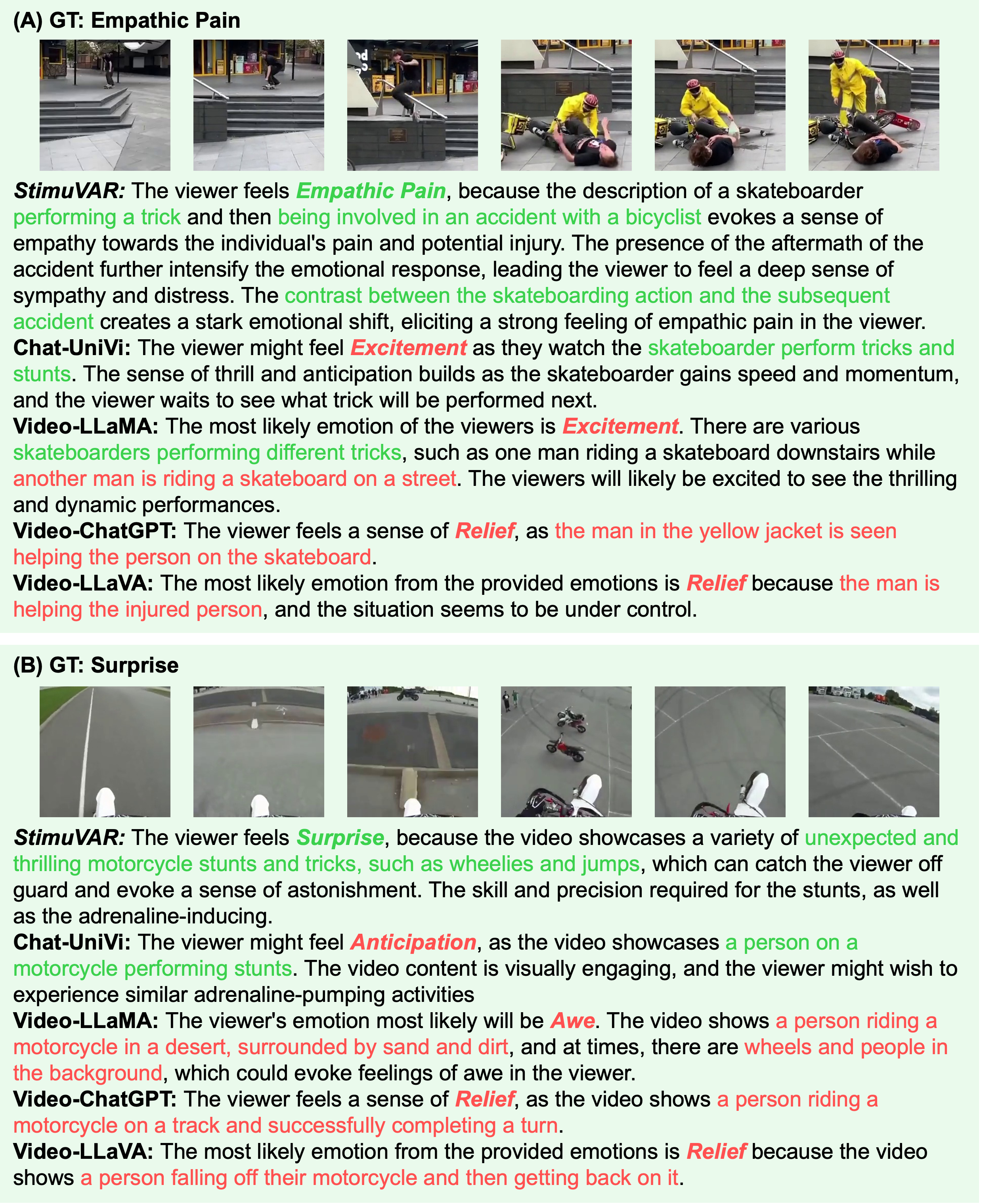}
    \caption{Output responses of different MLLMs. Compared to the baselines, StimuVAR accurately captures the scenes, activities and emotional stimuli, and then connects these elements with viewers' emotional reactions. \textcolor{relevant}{Relevant} and \textcolor{irrelevant}{irrelevant} words are colored.}
    \label{fig:examples}
\end{figure}

\noindent\textbf{Qualitative results.}
\cref{fig:examples} provides qualitative results of two examples. We can see that the responses generated by our StimuVAR successfully capture the key stimuli events (e.g., ``being involved in an accident with a bicyclist'' and ``unexpected and thrilling motorcycle stunts and tricks'') and offer causal connections between the stimuli and viewers' emotional reactions (e.g., ``accident → injury → sense of sympathy'' and ``motorcycle tricks → sense of astonishment''). In contrast, other MLLM baselines fail to capture the accidents that have a strong emotional intensity, resulting in incorrect predictions. This shows the importance of the proposed stimuli-aware mechanisms.

\subsection{Ablation Study}
\label{sec:ablation}
In this section, we look into how the proposed mechanisms affect StimuVAR. We conduct experiments on the VCE dataset~\cite{mazeika2022would}.

\begin{table}[!t]

\centering
\caption{Left: Ablation on the proposed strategy design. Right: Ablation on temporal aggregation design. \#Tokens denotes the number of spatial tokens. Results are on the VCE dataset. Both Top-1 and Top-3 accuracy are reported.}

\setlength{\tabcolsep}{13pt}
\begin{tabular}{l  l}

\begin{subtable}[h]{.45\linewidth}
\centering
\renewcommand{\arraystretch}{1.2} 
\setlength{\tabcolsep}{1pt}
    {\begin{tabular}{lrr}
    
    \toprule
        Modules & Top-1 & Top-3\\
        \midrule
        Vanilla model & 44.5 & 68.4\\
        $+$ event-driven frame sampling & 46.0 & 70.9\\
        $+$ VAR visual instruction tuning & 47.3 & 72.5\\
        $+$ emotion-triggered tube selection & \textbf{48.3} & \textbf{73.5}\\
       
        \bottomrule
    \end{tabular}}
\end{subtable}
\hspace{0.1in}

&

\begin{subtable}[h]{.38\linewidth}
\centering
\setlength{\tabcolsep}{1pt}
    {\begin{tabular}{lrrr}
    
    \toprule
        Modules & \#Tokens & Top-1 & Top-3\\
        \midrule
        Valley V1~\cite{luo2023valley} &256 & 44.5 & 68.4\\
        Valley V2~\cite{luo2023valley} &256& 46.6 & 71.2\\
        Valley V3~\cite{luo2023valley} &256& 47.5 & 72.9\\
        TS Selection~\cite{wang2022efficient} & 128 & 47.8 & 72.5\\
        StimuVAR (Ours) & 128 & \textbf{48.3} & \textbf{73.5}\\
        \bottomrule
    \end{tabular}}
\end{subtable}

\end{tabular}

\label{tab:ablation}
\end{table}

\begin{figure}[!t]
    \centering
    \includegraphics[width=1\linewidth]{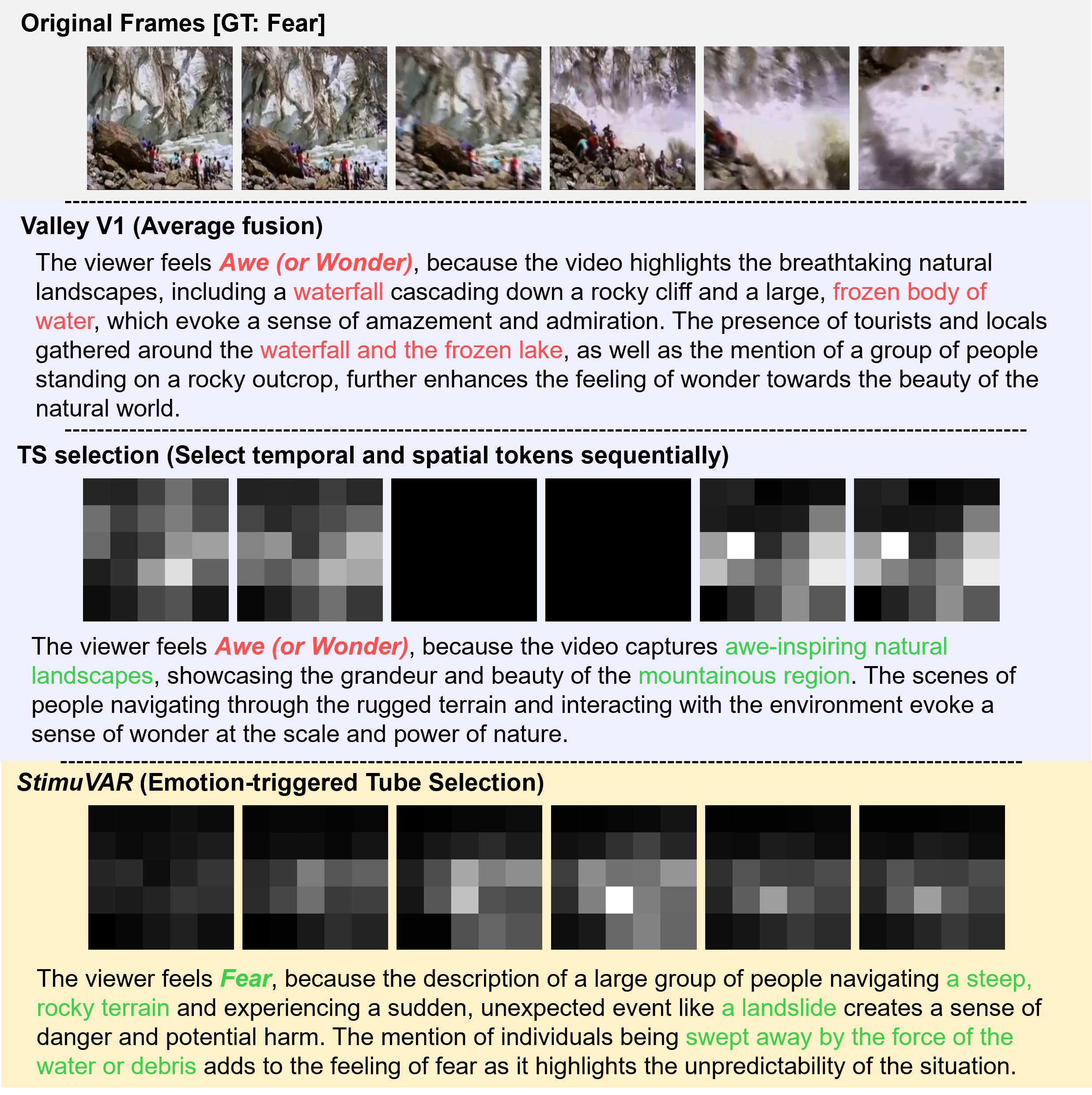}
    \caption{Comparison of tube score maps and output responses of different temporal aggregation designs. Valley~\cite{luo2023valley} often hallucinates objects that do not appear in a video. TS Selection~\cite{wang2022efficient} could be easily misled by its inaccurate temporal selection. In contrast, our emotion-triggered tube selection simultaneously considers both spatial and temporal information, resulting in the accurate selection of emotion-relevant tokens. \textcolor{relevant}{Relevant} and \textcolor{irrelevant}{irrelevant} words are colored.}
    \label{fig:temporal_module}
    \vspace{-5mm}
\end{figure}

\noindent \textbf{Strategy design.}
In this work, we propose three key components: event-driven frame sampling, emotion-triggered tube selection, and affective training. The ablation results are shown in \Cref{tab:ablation} (Left). We define a Vanilla model that utilizes the \texttt{<CLS>} token along with averaged \texttt{<Patch>} tokens to represent a video, and it is trained by only the initial phase of affective training. This baseline obtains 44.5\% Top-1 and 68.4\% Top-3 accuracy. Adding our event-driven frame sampling strategy improves Top-1/Top-3 accuracy by 1.5\%/2.5\%, demonstrating the effectiveness of localizing frame-level temporal stimuli. This positive trend continues with the inclusion of our VAR visual instruction tuning (i.e., the second phase of affective training), which increases Top-1/Top-3 accuracy by 1.3\%/1.6\%. The VAR visual instruction tuning enhances the connections between stimuli and viewers' emotions through the generated reasoning, achieving deeper affective understanding. Finally, further integrating the emotion-triggered tube selection module (i.e., the final StimuVAR, integrating all three components) achieves the highest performance, reaching 48.3\%/73.5\% Top-1/Top-3 accuracy. This improvement is due to the identification of spatiotemporal stimuli in the token space. These results demonstrate the effectiveness of all three proposed strategies.

\noindent \textbf{Temporal aggregation design.}
Due to the computational cost of attention increasing with increasing numbers of visual tokens, it is challenging for MLLMs to encode videos effectively and efficiently. Valley~\cite{luo2023valley} introduces three variants of temporal aggregation to encode video representations: (V1) averages the \texttt{<Patch>} tokens; (V2) computes the weighted sum of the \texttt{<Patch>} tokens; and (V3) combines the averaged \texttt{<Patch>} tokens with the dynamic information extracted by an additional transformer layer on the \texttt{<Patch>} tokens. Wang et al.~\cite{wang2022efficient} design a token selection module for efficient temporal aggregation, which considers temporal and spatial dimensions sequentially to select tokens; we name it TS Selection. The proposed StimuVAR uses emotion-triggered tube selection for temporal aggregation, considering spatiotemporal information simultaneously. In this experiment, we replace StimuVAR's tube selection with these baseline temporal aggregation designs to compare their effectiveness. As shown in \Cref{tab:ablation} (Right), StimuVAR's emotion-triggered tube selection outperforms Valley V1 - V3. This is because Valley V1 - V3 fuse spatial features in the temporal dimension, which is prone to producing objects not appearing in the video in the feature domain, causing visual hallucinations. StimuVAR also outperforms TS selection, demonstrating the superiority of the spatiotemporal design of our tube selection. \cref{fig:temporal_module} illustrates the hallucination issue of Valley V1 - V3. Specifically, the fused features incorrectly produced objects like `\textit{fozen body of water}', `\textit{frozen lake}' and `\textit{waterfall}', which are not present in the video. On the other hand, TS Selection relies heavily on temporal selection. As shown in the middle of  \cref{fig:temporal_module}, if frames containing emotion-related events are mistakenly filtered out by temporal selection, the final output would be misled. In contrast, StimuVAR uses tubes as the minimal unit of selection, mitigating such issues. Our method achieves higher performance while reducing the number of tokens to enhance efficiency.

\noindent\textbf{Fine-tune state-of-the-art video-oriented MLLMs using VAR instruction data.}
To exclusively assess the contributions of our proposed VAR instruction data, we use our VAR instruction data to fine-tune the three best-performing MLLMs in \Cref{tab:VCE_main}: Chat-UniVi~\cite{Jin_2024_CVPR}, Valley~\cite{luo2023valley}, and VideoChat2~\cite{li2024mvbench}, where we follow their official fine-tuning guidelines for each model. We evaluate them on the VCE, VE-8 and YF-6 datasets, using the same evaluation metrics described in \cref{dataset}. As can be seen in \Cref{tab:finetune}, fine-tuning using our VAR instruction data consistently improves the performance of all three MLLMs on all three datasets. This highlights the quality and effectiveness of the proposed VAR instruction data in enhancing MLLMs' affective reasoning capabilities. On the other hand, compared to these fine-tuned MLLMs, our StimuVAR still consistently demonstrates clear superiority. This highlights the advantages of the proposed model design. Overall, these results demonstrate that both the proposed VAR instruction data and model design are effective, playing their roles in contributing to excellent performance.

\renewcommand{\arraystretch}{0.9}
\begin{table}[t!]
    \centering
    \setlength{\tabcolsep}{3.5pt}
    \caption{Quantitative comparison of state-of-the-art video-oriented MLLMs, with and without fine-tuned on our proposed VAR instruction data.}
    \label{tab:finetune}
    \begin{tabular}{l |r | r |  rrrr | r | rrrr}
    \toprule
       \phantom{+}Methods & Acc & Emo-ali & RR & RW & WR & WW & CLIP-S & Win & Lose & Tie &Score\\
\midrule

       \small \textbf{VCE} & & & & & & & & & & & \\
       \phantom{+}Valley 
       {\includegraphics[width=0.3cm]{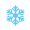}}~\cite{luo2023valley}&  31.3 & 29.4 & 19.2 & 12.1 & 10.2 & 58.5 & 69.4 & 45 & 12 & 43 & 2.89\\ 
       \phantom{+}Valley \includegraphics[width=0.3cm]{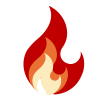}~\cite{luo2023valley}  & 42.1& 43.0& 38.4 & 3.7 & 4.6 & 53.3 & 70.5 &32 &14 &54 & 3.21 \\ 
       \phantom{+}VideoChat2
       {\includegraphics[width=0.3cm]{ice.png}}~\cite{li2024mvbench} & 31.1 & 36.4 & 24.0 & 7.1 & 12.4 & 56.5 & 68.6 & 56 & 12 & 32 & 2.65\\
       \phantom{+}VideoChat2\includegraphics[width=0.3cm]{Flame.png}~\cite{li2024mvbench}  & 40.1 & 40.2& 28.6 &11.5 & 11.6& 48.3 & 69.4 & 39&16 &45 & 3.15\\
       \phantom{+}Chat-UniVi
       {\includegraphics[width=0.3cm]{ice.png}}~\cite{Jin_2024_CVPR} &  38.6 &29.5 & 21.0 & 17.6 & 8.5 & 52.9  & 70.2& 50 & 19 & 31 & 2.83\\
       \phantom{+}Chat-UniVi\includegraphics[width=0.3cm]{Flame.png}~\cite{Jin_2024_CVPR}  & 56.8 & 48.5 & 45.3& 11.5 & 3.2 & 40.0 & 71.8 & 36&15 &48 & 3.22\\
       
       \textbf{StimuVAR (Ours)} & \textbf{73.5} & \textbf{69.6} & \textbf{68.8} & 4.7 & 0.8 & 25.6 & \textbf{75.3}& - & - & - & \textbf{3.40}\\
\midrule [1pt]

       \small \textbf{VE-8} & & & & & & & & & & & \\
       \phantom{+}Valley 
       {\includegraphics[width=0.3cm]{ice.png}}~\cite{luo2023valley}&  16.1  & 20.6 & 12.2 & 4.0 & 8.5 & 75.3 & 67.0& 35 & 7 & 58 & 2.88\\ 
       \phantom{+}Valley \includegraphics[width=0.3cm]{Flame.png}~\cite{luo2023valley}  & 25.5 & 26.6 &21.1 & 4.4& 5.5 & 69.0& 68.4& 33 & 17&50 & 3.05\\ 
       \phantom{+}VideoChat2
       {\includegraphics[width=0.3cm]{ice.png}}~\cite{li2024mvbench} & 16.5  & 15.6 & 14.3 & 2.2 & 1.4 & 82.1 & 65.1& 48 & 13 & 39 & 3.02\\
       \phantom{+}VideoChat2\includegraphics[width=0.3cm]{Flame.png}~\cite{li2024mvbench} & 32.7 & 29.9 & 27.1 & 5.6 & 2.8& 64.5& 67.9 & 27 & 22& 51& 3.12\\
       \phantom{+}Chat-UniVi
       {\includegraphics[width=0.3cm]{ice.png}}~\cite{Jin_2024_CVPR} &  14.4  &19.2 & 9.9 & 4.5 & 9.3 & 76.3  & 65.9 & 37 & 20 & 43 & 3.08\\
       \phantom{+}Chat-UniVi\includegraphics[width=0.3cm]{Flame.png}~\cite{Jin_2024_CVPR} & 25.4 & 28.6 & 20.1 & 5.3& 8.5& 66.1 & 66.3& 26& 19 & 55 & 3.15\\
       
       \textbf{StimuVAR (Ours)} & \textbf{40.5}  & \textbf{41.7} & \textbf{33.7} & 6.8 & 8.0 & 51.5 & \textbf{70.6}& - & - & - & \textbf{3.29}\\
\midrule [1pt]
       \small \textbf{YF-6} & & & & & & & & & & & \\
       \phantom{+}Valley 
       {\includegraphics[width=0.3cm]{ice.png}}~\cite{luo2023valley}&  10.0  & 18.9 & 5.9 & 4.1 & 13.0 & 77.0 & 66.4& 44 & 10 & 46 & 2.89\\ 
       \phantom{+}Valley \includegraphics[width=0.3cm]{Flame.png}~\cite{luo2023valley}  &  18.0& 22.0& 12.6& 5.4& 9.4& 72.6& 66.8& 28& 19& 53& 3.14\\ 
       \phantom{+}VideoChat2
       {\includegraphics[width=0.3cm]{ice.png}}~\cite{li2024mvbench} & 24.3  & 23.7& 18.5 & 5.8 & 5.2 & 70.5 & 64.1& 37 & 16 & 47 & 3.03\\
       \phantom{+}VideoChat2\includegraphics[width=0.3cm]{Flame.png}~\cite{li2024mvbench}  & 39.7&35.6 &34.1 & 5.6 & 1.5& 58.8& 65.1& 30& 21& 49& 3.16\\
       \phantom{+}Chat-UniVi
       {\includegraphics[width=0.3cm]{ice.png}}~\cite{Jin_2024_CVPR} &  28.4  &29.2 & 22.1 & 6.3 & 7.1 & 64.5  & 66.3& 33 & 14 & 53 & 3.12\\
       \phantom{+}Chat-UniVi\includegraphics[width=0.3cm]{Flame.png}~\cite{Jin_2024_CVPR}  & 32.7&  30.3& 22.8& 9.9&7.5 & 59.8 &  67.0 &26 & 22& 52& 3.18\\
       \textbf{StimuVAR (Ours)} & \textbf{46.2}  & \textbf{44.2} & \textbf{37.3} & 8.9 & 6.9 & 46.9 & \textbf{72.1}& - & - & - & \textbf{3.30}\\
    \bottomrule
    \end{tabular}
    \label{tab:finetune}
\end{table}

\color{black}

\begin{figure}[!t]
    \centering
    
        \begin{subfigure}[c]{.49\linewidth}
            \includegraphics[width=.9\textwidth, height=.6\textwidth]{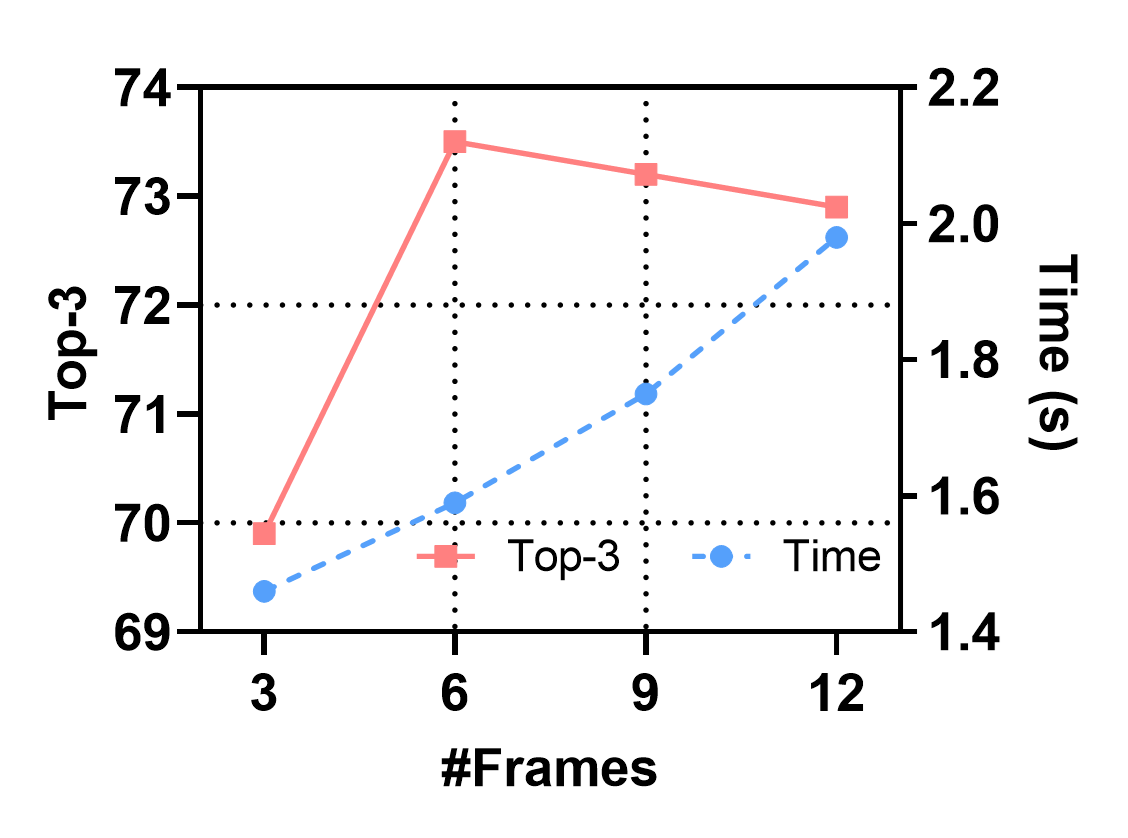}            
            \label{fig:frames}
        \end{subfigure}
        \begin{subfigure}[c]{.49\linewidth}
            \includegraphics[width=.9\textwidth, height=.6\textwidth]{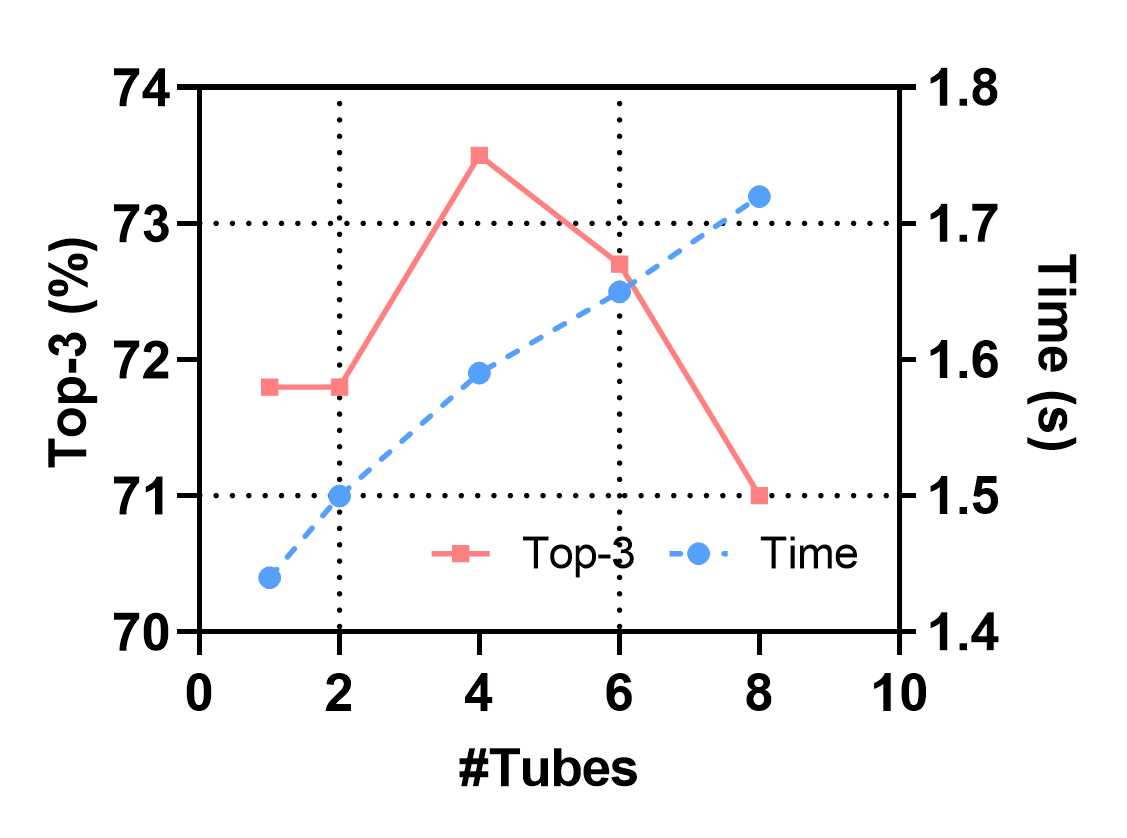}
            \label{fig:tube}
        \end{subfigure} \\
    \vspace{-2mm}
    \caption{Ablation on the number of selected frames (left) and tubes (right) for accuracy and inference time on the VCE dataset.}
   \label{fig:temporal_tube}
   \vspace{-4mm}
\end{figure}

\noindent \textbf{Number of selections in stimuli awareness.}
We explore varying numbers of frames to sample and tubes to select in our event-driven frame sampling and emotion-triggered tube selection strategies, respectively. As shown in \cref{fig:temporal_tube}, initially increasing the numbers of selected frames $N$ and tubes $K$ improves performance. Specifically, the Top-3 accuracy increases from 69.9\% to 73.5\% with three more frames (from $N=3$ to $N=6$), and there is an improvement from 71.8\% to 73.5\% when we select the top $K=4$ tubes compared to $K=1$. These increases are attributed to the additional information available, which allows the model to gain a more comprehensive affective understanding of the videos. However, when we continue to increase the number of frames and tubes, the performance begins to decline. The Top-3 accuracy decreases to 72.9\% when $N=12$ frames are sampled, and it also drops to 71.0\% when we double the tubes to $K=8$. This decline is due to redundant information from excessive frames or tokens, which distracts the model from emotional stimuli. Still, StimuVAR consistently outperforms the MLLM baselines (as presented in \Cref{tab:VCE_main}) with these different $N$ and $K$, demonstrating that the proposed strategies are not very sensitive to hyperparameter choices or computational budgets. 
Regarding computational cost, as expected, increasing the number of selected frames $N$ and tubes $K$ results in longer inference times (about linearly). This highlights the importance and effectiveness of the proposed event-driven frame sampling and emotion-triggered tube selection strategies, which can improve both accuracy and computational efficiency.


\renewcommand{\arraystretch}{0.9}
\begin{table}[t!]
    \centering
    \setlength{\tabcolsep}{5.5pt}
    \caption{Ablation on the property of event-driven frame sampling on the VCE dataset. Top-3 accuracy is reported.}
    \label{tab:Suprise}
        \begin{tabular}{l |ccc |ccc}
            \toprule
            Test set & \multicolumn{3}{c!{\vrule width 0.5pt}}{Surprise subset} & \multicolumn{3}{c!}{Entire test set} \\
            \midrule
            Frame sampling & Uniform & AKS~\cite{tang2025adaptive} &Event-driven & Uniform & AKS~\cite{tang2025adaptive} &Event-driven\\
            \midrule            VideoChat2{\includegraphics[width=0.3cm]{ice.png}}~\cite{li2024mvbench} & 13.7 & 14.1 & \textbf{14.5} & 31.1& 31.8& \textbf{32.5}\\            VideoChat2{\includegraphics[width=0.3cm]{Flame.png}}~\cite{li2024mvbench} &14.8& 14.9 & \textbf{18.8} & 40.1& 41.2 & \textbf{43.4}\\
            Chat-UniVi{\includegraphics[width=0.3cm]{ice.png}}~\cite{Jin_2024_CVPR} & 18.1 & 18.3 & \textbf{25.7} &38.6 & 38.9 & \textbf{41.0}\\
            Chat-UniVi{\includegraphics[width=0.3cm]{Flame.png}}~\cite{Jin_2024_CVPR} & 57.9 & 57.8 & \textbf{63.2} &   56.8&  57.4 & \textbf{61.7}\\
            StimuVAR (Ours) & 67.6 & 69.3 & \textbf{72.9} &69.7 & 71.6 & \textbf{73.5}\\
             \bottomrule
        \end{tabular}
\end{table}

\noindent \textbf{Property of event-driven frame sampling.}
As discussed in \cref{Event}, the motivation behind proposing the event-driven frame sampling strategy is to effectively capture rapid, unexpected events that most likely evoke viewers' emotional reactions. \Cref{tab:ablation} (Left) validates the effectiveness of this strategy. In this experiment, we further analyze the properties of this strategy. Specifically, we notice that \textit{surprise} is one of the emotions particularly sensitive to rapid, unexpected events. In other words, such events inherently tend to evoke an emotion of surprise. To explore this, we examine various models' performance on the \textit{surprise} test subset of the VCE dataset, which consists of 2,529 videos annotated as \textit{surprise}.
In addition to the uniform sampling baseline, we further compare with AKS~\cite{tang2025adaptive}, a state-of-the-art frame selection method designed for long video understanding tasks. In \Cref{tab:Suprise}, given the same number of frames to select (i.e., $N=6$), event-driven sampling outperforms uniform sampling and AKS on all the models, where the improvements are even more significant on the surprise subset.
These results again demonstrate the effectiveness of event-driven frame sampling in capturing rapid, unexpected events as emotional stimuli at the frame level, leading to higher performance, especially for emotions sensitive to such events.

\begin{figure}[!t]
    \centering
    \includegraphics[width=1\linewidth]{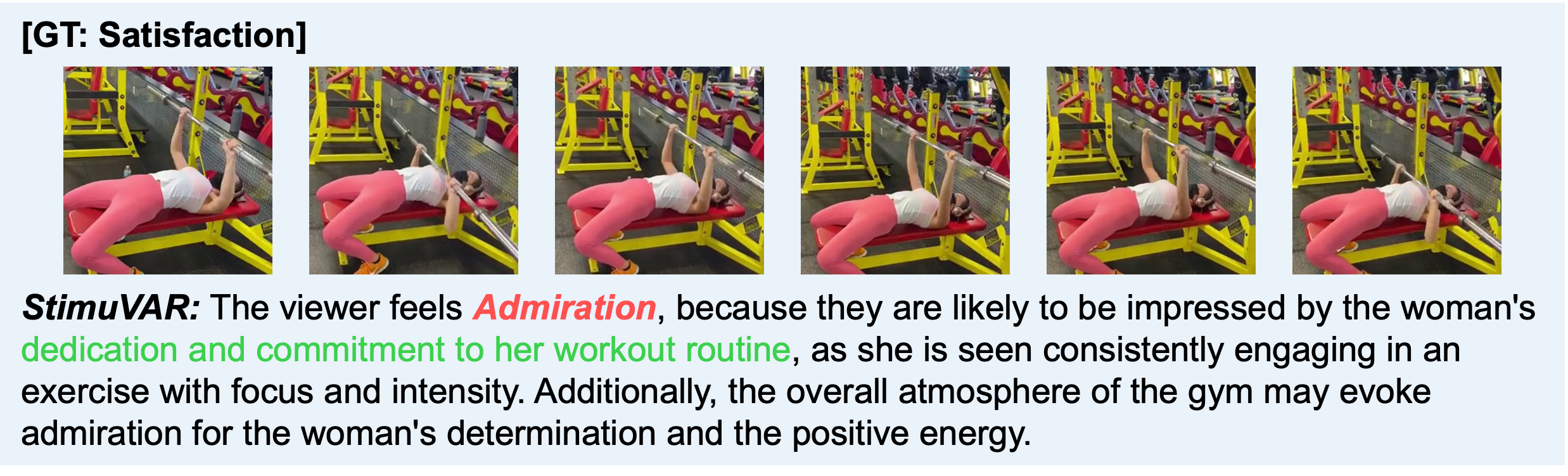}
    \caption{A failure case due to subjectivity. Although StimuVAR's emotional prediction (``admiration'') differs from the ground-truth label (``satisfaction''), it gives a plausible explanation. This ``incorrect'' prediction arises from the inherent subjectivity of emotion tasks, emphasizing the importance of incorporating reasoning.}
    \label{fig:failure}
    \vspace{-5mm}
\end{figure}

\noindent \textbf{Failure cases due to subjectivity.}
Emotion prediction is a task with an extent of subjectivity. Although humans share certain commonsense knowledge, the same visual stimuli may still evoke different emotional responses in different viewers. We observe this interesting phenomenon from our results as well, where StimuVAR makes predictions that differ from the ground-truth labels, yet its rationales are reasonable. For example, in \cref{fig:failure}, StimuVAR correctly describes a woman working out in a gym. The woman's dedication and commitment could evoke emotions of the predicted admiration, the ground-truth satisfaction, or both. These emotions are all reasonable. This example highlights the crucial role of reasoning, especially given the subjective nature of emotion tasks, as reasoning explains how the model makes its predictions and, consequently, how these results can be applied.



\begin{table}[t!]
\centering
\caption{Ablation on the generalizability across different LLMs on the VCE dataset.}
\setlength{\tabcolsep}{22pt}
    \begin{tabular}{c |c c c }
    \toprule
        LLM & Llama2-7b~\cite{touvron2023llama} & Llama2-13b~\cite{touvron2023llama} & Llama3-8b~\cite{dubey2024llama} \\
        \midrule
        Top-3 & 73.5 & 73.6 & 74.0\\
         \bottomrule
    \end{tabular}
\label{tab:llms}
\end{table}

\noindent \textbf{Generalizability across different LLMs.}
We apply StimuVAR to the MLLM backbone with different LLMs. \Cref{tab:llms} shows that increasing the model size leads to performance improvement. Using the more advanced LLM, Llama3-8b~\cite{dubey2024llama}, further improves performance to 74.0\%. These results demonstrate that the proposed strategies have high generalizability across different backbone architectures and can consistently scale with larger/better backbones.

\section{Conclusion}\label{Conclu}

In this work, we propose StimuVAR, an MLLM-based spatiotemporal stimuli-aware framework for VAR. The event-driven frame sampling strategy employs optical flow as a cue to sample frames of key events at the frame level, while the emotion-triggered tube selection strategy further localizes the emotional stimuli areas at the token level. Additionally, we create VAR visual instruction data used for affective training, enhancing StimuVAR's affective understanding and reasoning ability. To the best of our knowledge, StimuVAR is the first MLLM-based method for VAR from the viewers' perspective. Extensive experiments demonstrate that StimuVAR effectively and efficiently predicts viewers' emotional reactions while also reasoning about its predictions. This dual capability enhances the model's reliability, making it suitable for real-world applications.

\noindent\textbf{Data availability statements.} 
All training/testing data and code that support the
findings of this study have been deposited in~\href{https://drive.google.com/drive/folders/1sRKitbXpLZ4pwXTONjiA-X0Y1z4I2o4X}{VCE},~\href{https://drive.google.com/drive/folders/0B5peJ1MHnIWGd3pFbzMyTG5BSGs?resourcekey=0-hZ1jo5t1hIauRpYhYIvWYA}{VE-8}, ~\href{https://drive.google.com/drive/folders/11uJTbqdHXqjQw63_teXXda0cILGDVyo4?usp=sharing}{YF-6} and~\href{https://www.dropbox.com/scl/fi/myue506itjfc06m7svdw6/EmoSet-118K.zip?rlkey=7f3oyjkr6zyndf0gau7t140rv&e=1&dl=0}{EmoSet}.

\newpage
\bibliography{main}

\newpage
\appendix

\renewcommand{\thefigure}{\Alph{figure}}
\renewcommand{\thetable}{\Alph{table}}
\setcounter{figure}{0}
\setcounter{table}{0}



\section{More Ablation Study}

\subsection{Claude 3.5 as a Judge}
To have a comprehensive evaluation using ``LLM-as-a-judge'', we employ Claude 3.5~\cite{anthropic2024claude} as an alternative judge, considering both its capabilities and cost-effectiveness. Similar to the approach used with GPT-3.5~\cite{brown2020language}, we assign scores as integers within the range of $[1, 4]$, and report 1-vs.-1 (StimuVAR vs. a baseline) comparison along with average scores on the same 100 test samples. As shown in \Cref{tab:claude}, while the average scores estimated by Claude 3.5 differ slightly from those produced by GPT-3.5, the ranking of reasoning quality across different methods remains consistent. This consistency, evaluated by two powerful LLMs, shows that StimuVAR demonstrates high reasoning quality.

\begin{table}[htb]
    \centering
    \setlength{\tabcolsep}{4pt}
    \caption{LLM-as-a-judge comparison on the VCE, VE-8 and YF-6 datasets using Claude 3.5. 100 test videos are randomly selected for evaluation. ``Win/Lose/Tie'' denotes the number of samples that StimuVAR wins/loses/ties against each competitor in a 1-vs.-1 comparison. ``Score'' denotes the average score on the 100 test samples.}
    \begin{tabular}{l | rrrr | rrrr | rrrr}
    \toprule
        Datasets & \multicolumn{4}{c !{\vrule width 1pt}}{VCE~\cite{mazeika2022would}} & \multicolumn{4}{c!{\vrule width 1pt}}{VE-8~\cite{jiang2014predicting}} &\multicolumn{4}{c}{YF-6~\cite{xu2016heterogeneous}}\\
        \midrule
        Judgement & Win & Lose & Tie & Score & Win & Lose & Tie & Score & Win & Lose & Tie & Score\\
        \midrule
        Video-LLaMA~\cite{Zhang2023VideoLLaMAAI} & 41 & 16 & 43 & 2.94 & 53 & 14 & 33 & 2.21 &57 & 11 & 32 & 2.16\\
        Video-LLaVA~\cite{Lin2023VideoLLaVALU} & 38 & 20 &42 & 2.88 & 58 & 13 & 29 & 2.13 & 59 & 22 & 19 & 2.36\\
        Valley~\cite{luo2023valley} & 47 & 18 & 35 & 2.84 & 45 & 18 & 37 & 2.72& 52 & 17 & 31 & 2.55\\
        Video-ChatGPT~\cite{Maaz2023VideoChatGPTTD} & 50 & 14 & 36 & 2.54 & 35 & 20 & 45 & 3.15 & 42 & 24 & 34 & 3.05\\
        VideoChat2~\cite{li2024mvbench} & 56 & 11 & 33 & 2.21 & 40 & 20 & 40 & 2.95 & 42 & 22 & 32 & 3.06\\
        Chat-UniVi~\cite{Jin_2024_CVPR} & 52 & 15 & 33 & 2.55 & 37 & 21 & 42 & 3.05 & 37 & 21 & 52 & 3.12\\
        mPLUG-Owl~\cite{Ye_2024_CVPR} & 26 & 17 & 57 & 3.29 & 36 & 22 & 42 & 3.12 & 39 & 22& 39 & 3.10\\
        EmoVIT~\cite{xie2024emovit} & 79 & 1 & 20 & 1.19 & 80 & 0 & 20 & 1.00  & 82 & 0 & 18 & 1.00\\
        \midrule
        StimuVAR (Ours) & - & - & - & \textbf{3.40} & - & - & - & \textbf{3.34} & - & - & - & \textbf{3.32}\\
    \bottomrule
    \end{tabular}
    \label{tab:claude}
\end{table}

\subsection{Comparison with Sophisticated LLMs}
To compare the performance with sophisticated LLMs, we explore directly input the sampled frames in to sophisticated MLLMs, e.g. GPT-4V~\cite{achiam2023gpt} and Claude 3.5~\cite{anthropic2024claude}. Because of their expensive inference cost, in this experiment, we randomly select 100 videos from the test set of VCE for evaluation. The results are shown in the \Cref{tab:visionmodel} below. We can see that StimuVAR significantly outperforms these models.

\begin{table}[t!]
    \centering
    \caption{Comparison with sophisticated vision models on VCE dataset.}
    \setlength{\tabcolsep}{15pt}
    \begin{tabular}{c|ccc}
    \toprule
        Model & GPT-4V~\cite{achiam2023gpt} & Claude 3.5 sonnet~\cite{anthropic2024claude} & StimuVAR\\
    \midrule
        Accuracy & 35 & 33 & 73\\
    \bottomrule
    \end{tabular}
    
    \label{tab:visionmodel}
\end{table}

On the other hand, as described in Sec.4.1, our paper has considered comprehensive evaluation metrics, but these metrics are reference-free since existing viewer-centered VEA datasets~\cite{mazeika2022would,jiang2014predicting,xu2016heterogeneous} do not contain ground-truth reasoning.
To make our evaluation more thorough, here we generate ``ground-truth'' via GPT and use the sentence semantic similarity~\cite{1644735} as an additional reference-based metric.
Specifically, we follow the same methodology (Sec.3.2) used to construct our VAR instruction dataset (based on the VCE training set) to generate GPT-based reasoning on the VCE test set. To measure the semantic similarity between model predictions and this reference set, we employ the sentence semantic similarity metric~\cite{1644735}, using a sentence encoder to extract and compare feature representations of model predictions and ``ground-truth reasoning". As shown in \Cref{tab:gpt_reference}, StimuVAR's responses exhibit closer semantic meaning and content alignment with the GPT-based reference reasoning. It outperforms all the baseline MLLMs regardless of whether they are fine-tuned on our VAR instruction data.

\begin{table}[t!]
    \centering
    \caption{Evaluation with GPT-based reference reasoning.}
    \setlength{\tabcolsep}{40pt}
    \begin{tabular}{l|r}
    \toprule
        Models & Similarity \\
    \midrule
        Video-LLaMA \raisebox{-0.15cm} {\includegraphics[width=0.5cm]{ice.png}} & 0.459 \\
        Video-LLaVA \includegraphics[width=0.4cm]{Flame.png} & 0.465 \\
        Valley \raisebox{-0.15cm} {\includegraphics[width=0.5cm]{ice.png}} & 0.492 \\
        Valley \includegraphics[width=0.4cm]{Flame.png} & 0.517 \\
        VideoChat2 \raisebox{-0.15cm} {\includegraphics[width=0.5cm]{ice.png}} & 0.489 \\
        VideoChat2 \includegraphics[width=0.4cm]{Flame.png} & 0.532 \\
        Chat-UniVi \raisebox{-0.15cm} {\includegraphics[width=0.5cm]{ice.png}} & 0.460 \\
        Chat-UniVi \includegraphics[width=0.4cm]{Flame.png} & 0.549 \\
        StimuVAR & \textbf{0.607} \\
    \bottomrule
    \end{tabular}
    
    \label{tab:gpt_reference}
\end{table}

We also design a direct baseline, which recognizes visual content and then leverages an external LLM to integrate commonsense knowledge of human reactions and emotions to reason the viewer's reaction. To implement this, we follow a process similar to that used in constructing the VAR instruction data. Specifically, we use a video-language model~\cite{wang2023cogvlm} to generate captions for each sampled frame, organizing them in a structured format and GPT then processes these frame-level captions, taking temporal correlations into account, to produce a coherent video-level caption. Subsequently, we leverage two closed-source models, Claude 3.5~\cite{anthropic2024claude} and GPT-3.5~\cite{brown2020language}, as well as an open-source model, Llama3-8b~\cite{dubey2024llama}, to infer the viewer’s likely emotion and generate the corresponding reasoning. As shown in~\Cref{tab:caption2feel}, StimuVAR outperforms these baselines by more than 50\% even though these are sophisticated LLMs with detailed visual content as input. The quality of image and video captions is another aspect to affect VAR performance. To further investigate this, we replace the initial captioning model, CogVLM~\cite{wang2023cogvlm}, with a more advanced VLM model, Qwen2.5-VL~\cite{bai2025qwen2}, to generate frame-level captions. While this upgrade does benefit downstream emotion recognition, such as 10.4\% improvement on Llama3-8b. Still, there remains a performance gap of over 40\% between these baselines and StimuVAR.
These results highlight the limitations of existing models in affective understanding, as well as the importance visual cues critical for recognizing emotion. 

\begin{table}[t!]
    \centering
    \caption{Comparison with sophisticated LLMs with video caption as input.}
    \setlength{\tabcolsep}{12pt}
    \begin{tabular}{c|ccc}
    \toprule
        Model & GPT-3.5~\cite{brown2020language} & Claude 3.5 sonnet~\cite{anthropic2024claude} & Llama3-8b~\cite{dubey2024llama}\\
    \midrule
                CogVLM~\cite{wang2023cogvlm} & 19.1 & 15.8 & 15.9 \\
        Qwen2.5-VL~\cite{bai2025qwen2} & 27.5 & 24.3 & 26.3 \\
    \bottomrule
    \end{tabular}
    
    \label{tab:caption2feel}
\end{table}

\subsection{Visualization of Emotion-triggered Tube Selection.}
\begin{figure}[!h]
    \centering
    \includegraphics[width=1\linewidth]{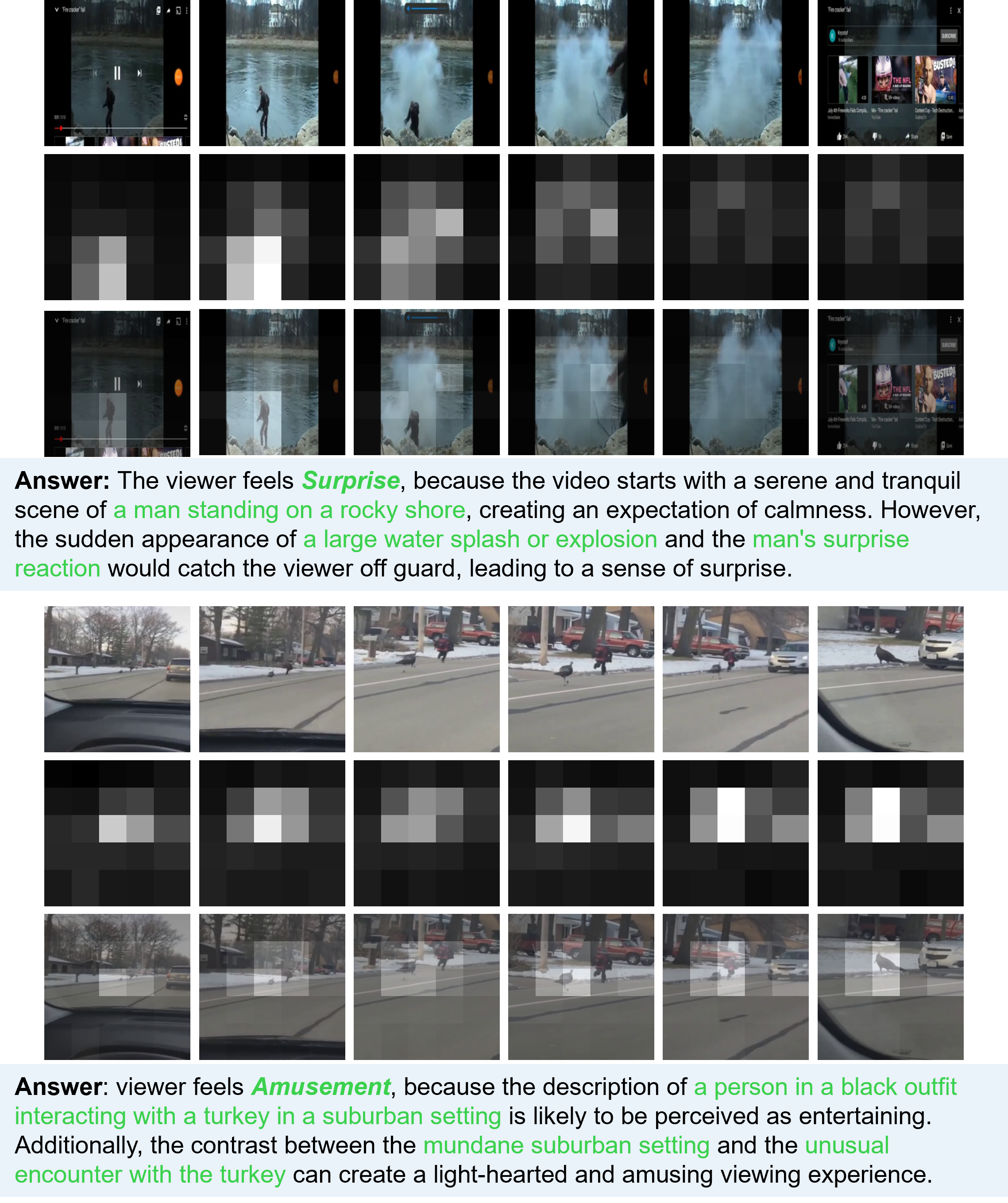}
    \caption{Visualization of StimuVAR's tube score maps and corresponding output responses. From top to bottom rows: original frames, tube score maps, and tube score maps overlaid on the frames. Emotion-triggered tube selection identifies the most emotion-relevant spatiotemporal areas, enabling effective and efficient video representations for VAR. \textcolor{relevant}{Relevant} and \textcolor{irrelevant}{irrelevant} words are colored.}
    \label{fig:tube_vis}
     \vspace{-5mm}
\end{figure}

As discussed, the emotion-triggered tube selection strategy enables StimuVAR to focus on spatiotemporal stimuli regions while reducing computational costs by using fewer tokens. In \cref{fig:tube_vis}, we present two more examples to visualize the token score maps. In the first example, higher scores are allocated to the spatiotemporal regions containing ``a man standing on a rocky shore and a large water splash or explosion'', while lower scores are given to regions with less emotionally relevant information, such as the last frame. In the second example, the primary focus is on an unexpected event in the video where a turkey is chasing a man in a black outfit. Most remaining areas do not explicitly trigger viewers' emotions and thus receive lower scores. Both examples demonstrate StimuVAR’s ability to select the essential tokens most likely representing emotional stimuli in the spatiotemporal domain. The visualization also shows a close alignment between the selected tubes and the keywords in the output explanations, highlighting how the proposed strategy benefits affective reasoning. Moreover, we observe that the emotional stimuli usually occupy only small portions of a video, indicating that many tokens are redundant to the VAR task, which explains StimuVAR’s high token efficiency.

\subsection{Extension to Image Data}
Although StimuVAR is designed to predict viewers' emotional reactions to videos, it can be extended to handle image inputs as well. This can be done by treating an image as a single-frame video. In this case, frame sampling is not useful, but token selection and affective training are still valid. We use EmoSet~\cite{yang2023emoset}, a large-scale emotion image dataset, to evaluate StimuVAR's performance on image data. EmoSet consists of 118,102 images manually labeled into one of eight emotion categories. In addition to prediction accuracy, we also measure the reasoning quality. To test StimuVAR's direct generalizability from videos to images, we evaluate the model trained with VCE data without fine-tuning on Emoset. We compare StimuVAR with image-oriented MLLMs~\cite{li2023blip,liu2024visual,Dai2023InstructBLIPTG} and EmoVIT~\cite{xie2024emovit}, where EmoVIT is explicitly fine-tuned on EmoSet. \Cref{tab:emoset} shows that StimuVAR outperforms image-oriented MLLMs in both prediction accuracy and reason quality by decent margins. Notably, while EmoVIT, fine-tuned on EmoSet, achieves the highest prediction accuracy, its reasoning ability is limited. In the LLM-as-a-judge evaluation, StimuVAR's reasoning wins in most samples, as EmoVIT fails to produce plausible reasoning consistently. These results demonstrate StimuVAR's deep understanding of human emotions and its generalizability from videos to images.

\begin{table}[t!]
    \centering
    \setlength{\tabcolsep}{4pt}
    \caption{Quantitative comparison on the EmoSet dataset. EmoVIT* indicates that EmoVIT is fine-tuned on the EmoSet data, while the other methods are not.}
    
    \begin{tabular}{l | r | r | rrrr | r | rrrr}
        \toprule
           Method  & Acc & Emo-ali & RR & RW & WR & WW & CLIP-S & Win & Lose & Tie & Score\\
        \midrule
           BLIP-2~\cite{li2023blip} & 46.8  & 43.2 & 36.5 & 10.3 & 6.7 & 46.5 & 65.5 & 98 & 0 & 2 & 1.26\\
           LLaVA~\cite{liu2024visual} & 44.0  & 40.1 &35.3 & 8.7 & 4.8 & 51.2 & 66.4 & 42 & 13 & 45 & 3.06\\
           InstructBLIP~\cite{Dai2023InstructBLIPTG} & 42.2  & 43.9 & 29.9 & 12.3 & 14.0 & 43.8 & 60.9 & 34 & 28 & 38 & 3.17\\
           EmoVIT*~\cite{xie2024emovit} & \textbf{83.4}  & 26.7 & 24.2 & 59.2 & 2.5 & 14.1 & 62.8 & 68 & 9 & 23 & 2.33\\       
        \midrule    
           StimuVAR (Ours) & 67.7 & \textbf{68.5} & \textbf{66.3} & 1.4 & 2.2& 30.1 & \textbf{70.5} & - & - & - & \textbf{3.34}\\
        \bottomrule
    \end{tabular} 
    \label{tab:emoset}
\end{table}

\section{Discussion}

\subsection{Frame Sampling Strategies in MLLMs}

In this work, we introduce event-driven frame sampling, which utilizes optical flow to perceptually localize plausible emotional stimuli at the frame level, enabling efficient video representation. In contrast, existing MLLMs, including Chat-UniVi\cite{Jin_2024_CVPR} and all other models listed in Table~1, rely on uniform frame sampling for video processing. 

Several strategies exist for extracting essential temporal information. For instance, Chat-UniVi~\cite{Jin_2024_CVPR} operates at the token level, employing visual token merging to aggregate dynamic visual tokens for image and video representation. However, it does not incorporate any adaptive frame selection beyond uniform sampling, making its research focus and methodology orthogonal to ours. 

While adaptive frame selection methods have been explored in traditional CNN-based models~\cite{wu2019adaframe,rahnama2022adaptive}, they fundamentally differ from our optical flow-based sampling and may not effectively capture emotional stimuli. For example, AdaFrame~\cite{wu2019adaframe} employs an LSTM network for frame selection, whereas our optical flow-based approach is significantly more efficient and better suited for affective reasoning tasks. By dynamically allocating higher frame rates to segments with significant perceptual changes, our method ensures that critical emotion-triggering moments are preserved while maintaining computational efficiency.

Alternative design approaches for long-video processing incorporate multiple or all video frames for representation. While processing all frames provides more comprehensive temporal coverage, it is impractical due to significant computational cost and temporal redundancy, even for models specifically designed for long videos~\cite{weng2024longvlm,buch2022revisiting}. In fact, these long-video models also rely on uniform frame sampling, despite employing techniques like token compression to reduce computational overhead. Their methods operate at the token or feature level only after a visual encoder has processed the uniformly sampled frames.

For example, LongVLM~\cite{weng2024longvlm} uniformly samples 100 frames from any input video, regardless of its duration (ranging from a few seconds to several minutes). After this sampling step, it applies token compression techniques. Our event-driven frame sampling strategy is orthogonal to such approaches, i.e., it can be integrated with them to enhance efficiency further. For instance, token compression can be applied to event-driven sampled frames instead of uniformly sampled ones.

As discussed, frame sampling is essential for existing models, as storing and processing visual tokens from all frames requires substantial GPU memory. To illustrate this computational burden, we conduct an analysis on Valley~\cite{luo2023valley}. As the number of input frames increases from 6 to 50, 100, and 200, GPU memory usage rises from 38.5 GB to 43.8 GB, 47.3 GB, and 54.4 GB, respectively, illustrating the scalability limitations. For a one-minute video (1,800 frames), the estimated GPU memory usage reaches 168 GB, and for a five-minute video (9,000 frames), memory consumption exceeds 679 GB. This substantial computational consumption severely constrains models from processing longer videos, making it infeasible for real-world applications.

In addition, capturing emotional stimuli is the main motivation of our event-driven frame sampling (see Fig.1 in the manuscript), as our work focuses on the affective reasoning problem. The event-driven frame sampling not only reduces computational cost but also enhances the capture of emotional stimuli. Since rapid, emotion-triggering events often occur within brief segments of a video, processing all frames can dilute attention to these critical moments. As shown in Table~5 (left), under the same sampling budget (selecting six frames; see Sec.4.1), event-driven frame sampling outperforms the Vanilla model with uniform sampling, demonstrating its effectiveness in capturing emotional stimuli.
Moreover, given the inherent temporal redundancy, allocating excessive computation to all frames is inefficient. By selecting frames at the input level before feature extraction, our method ensures that key segments receive prioritized attention in a more efficient manner.

Apart from VAR, the proposed event-driven sampling is also well-suited for tasks such as video anomaly detection, detecting the sudden appearance of bicycles or vehicles in pedestrian zones. It is also relevant to autonomous driving, where unexpected objects such as pedestrians or obstacles appear only in a few frames. Additionally, event-driven sampling is orthogonal to existing video understanding approaches relying on uniform sampling. For example, AKS~\cite{tang2025adaptive} uses similarity scores between uniformly sampled frames and question prompts, followed by a hierarchical judgment and splitting strategy. As shown in \Cref{tab:aks}, by replacing AKS's uniform sampling step with our event-driven sampling under 16 frames with CLIP score, we observe a 1.6\% improvement on the LongVideoBench~\cite{wu2024longvideobench} benchmark. This highlights the broader utility of the proposed frame selection method.

\begin{table}[t!]
    \centering
    \caption{AKS Performance on LongVideoBench with and without event-driven sampling.}
    \begin{tabular}{c|cc}
    \toprule
        LongVideoBench val & AKS~\cite{tang2025adaptive} & Event-driven + AKS\\
    \midrule
        Accuracy & 60.2 & 61.8\\
    \bottomrule
    \end{tabular}
    
    \label{tab:aks}
\end{table}

\subsection{VAR Instruction Data Details}
\begin{figure}[h!]
    \centering
    \includegraphics[width=\linewidth]{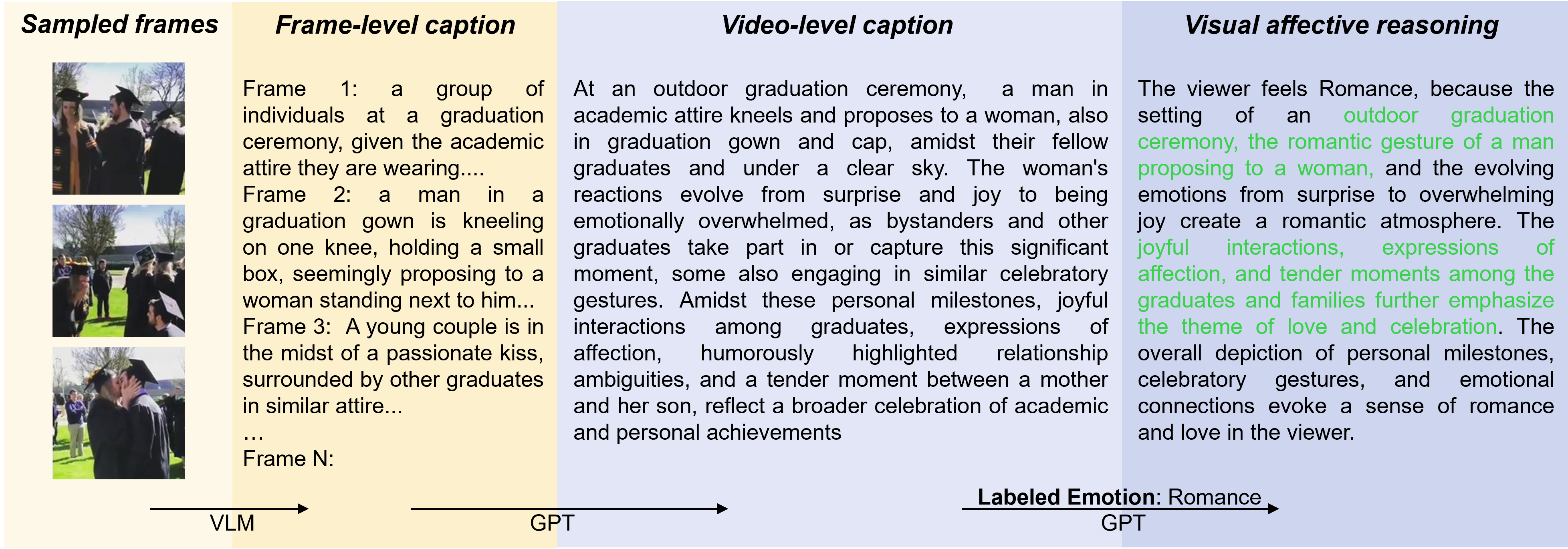}
    \caption{The process of generating visual affective reasoning.}
    \label{fig:generate}
\end{figure}

We draw~\cref{fig:generate} below to illustrate our instruction generation process more clearly, and also provide examples of VAR instruction data in~\cref{fig:var_examples}.

\subsection{User Study}

To measure the alignment of human preference and model-generated responses, we conduct a user study using 10 selected videos, involving 22 participants from around the world, including Baltimore, San Jose, Taipei, Hsinchu, Sydney, etc. We follow a similar evaluation setup to LLM-as-a-judge~\cite{zheng2024judging}, as described in \textit{Sec.4.1 Experimental Setup}. Specifically, for each video, we provide two anonymized, model-generated responses along with an option to indicate no preference. Participants are then asked to select the response they find better aligned with the video content, or choose no preference.

\begin{table}[t!]
    \centering
    \setlength{\tabcolsep}{20pt}
    \caption{User study comparison on the VCE dataset, where 10 videos are selected for evaluation. ``Win/Lose/Tie'' denotes the number of samples that StimuVAR wins/loses/ties against the competitor in a 1-vs.-1 comparison.}
    \begin{tabular}{l | rrr }
    \toprule
        Datasets & \multicolumn{3}{c !}{VCE~\cite{mazeika2022would}}\\
        \midrule
        Judgement & Win & Lose & Tie \\
        \midrule
       Chat-UniVi~\cite{Jin_2024_CVPR} & 8 & 0 & 2\\
    \bottomrule
    \end{tabular}
    
    \label{tab:user_study}

\end{table}

As shown in~\Cref{tab:user_study}, the results consistently favor StimuVAR, which is preferred in 8 out of 10 cases, while the remaining two received no preference. These findings further support the quality and interpretability of StimuVAR's affective reasoning.

\begin{figure}[h]
    \centering
    
    \begin{subfigure}[b]{\textwidth}
        \centering
        \includegraphics[width=\textwidth]{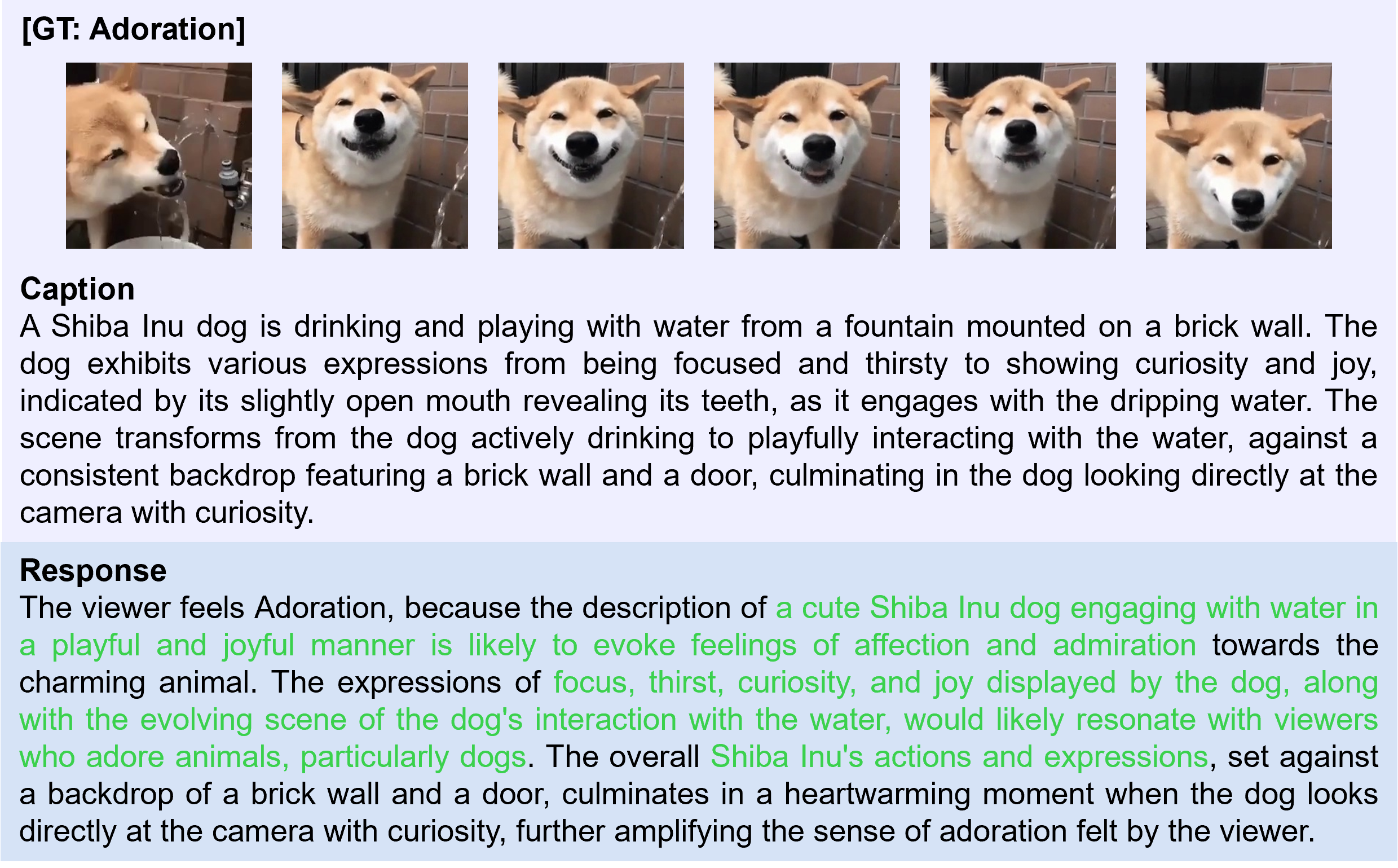} 
        \label{fig:sub1}
    \end{subfigure}
    \hfill
    \begin{subfigure}[b]{\textwidth}
        \centering
        \includegraphics[width=\textwidth]{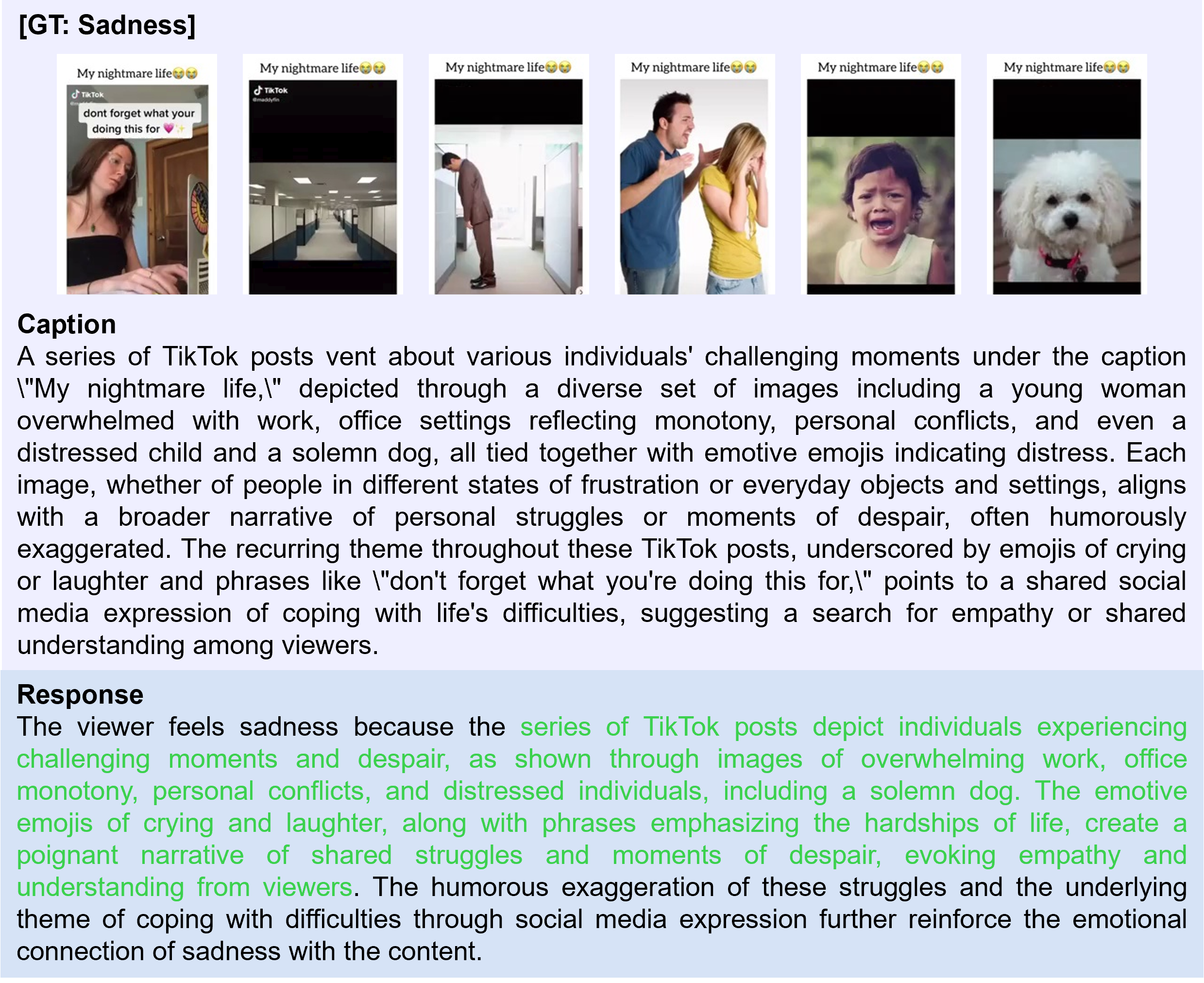} 
        \label{fig:sub2}
    \end{subfigure}
\end{figure}

\begin{figure}[h]
    \centering
    \begin{subfigure}[b]{\textwidth}
        \centering
        \includegraphics[width=\textwidth]{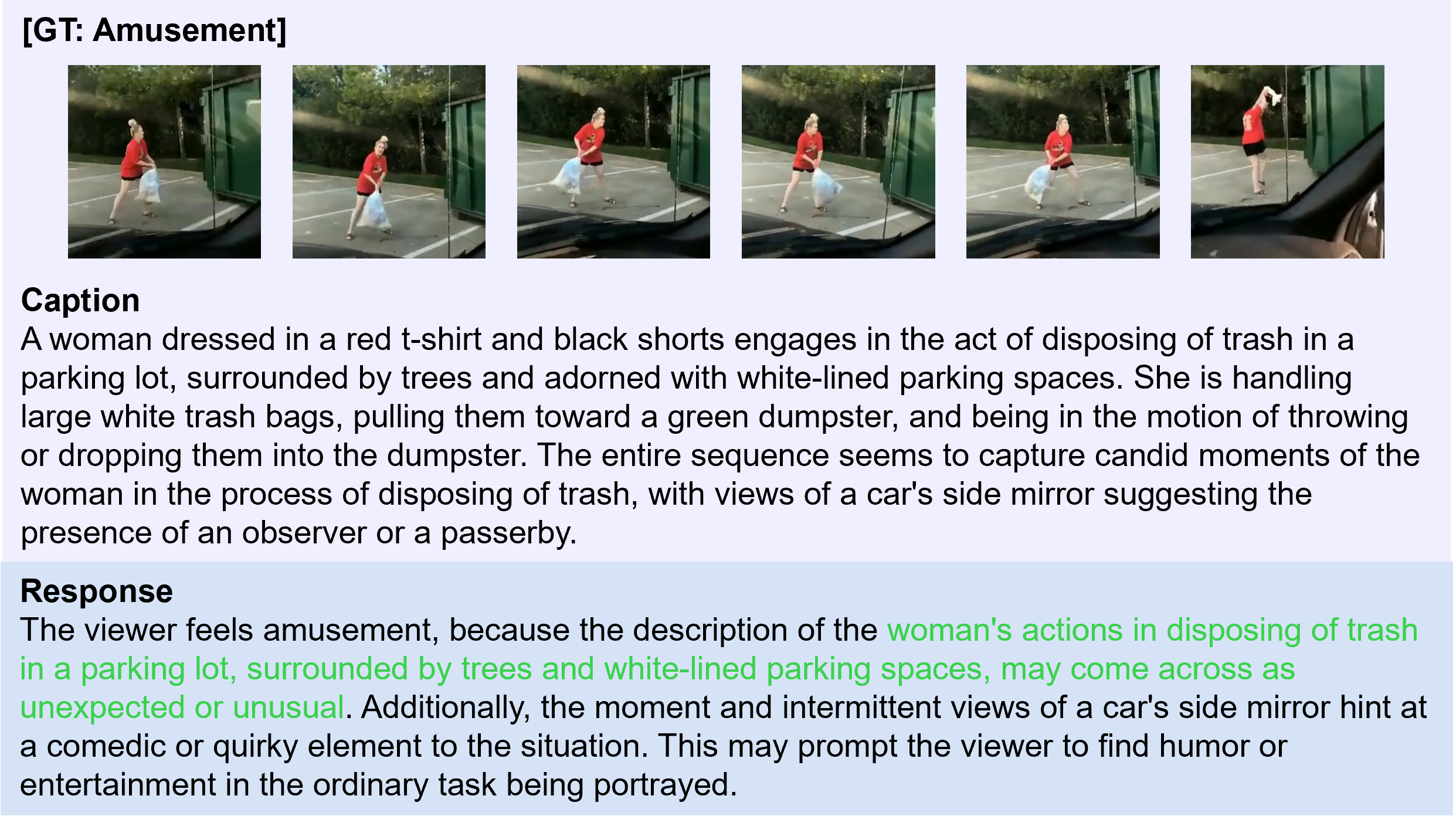} 
        \label{fig:sub3}
    \end{subfigure}
    \hfill
    \begin{subfigure}[b]{\textwidth}
        \centering
        \includegraphics[width=\textwidth]{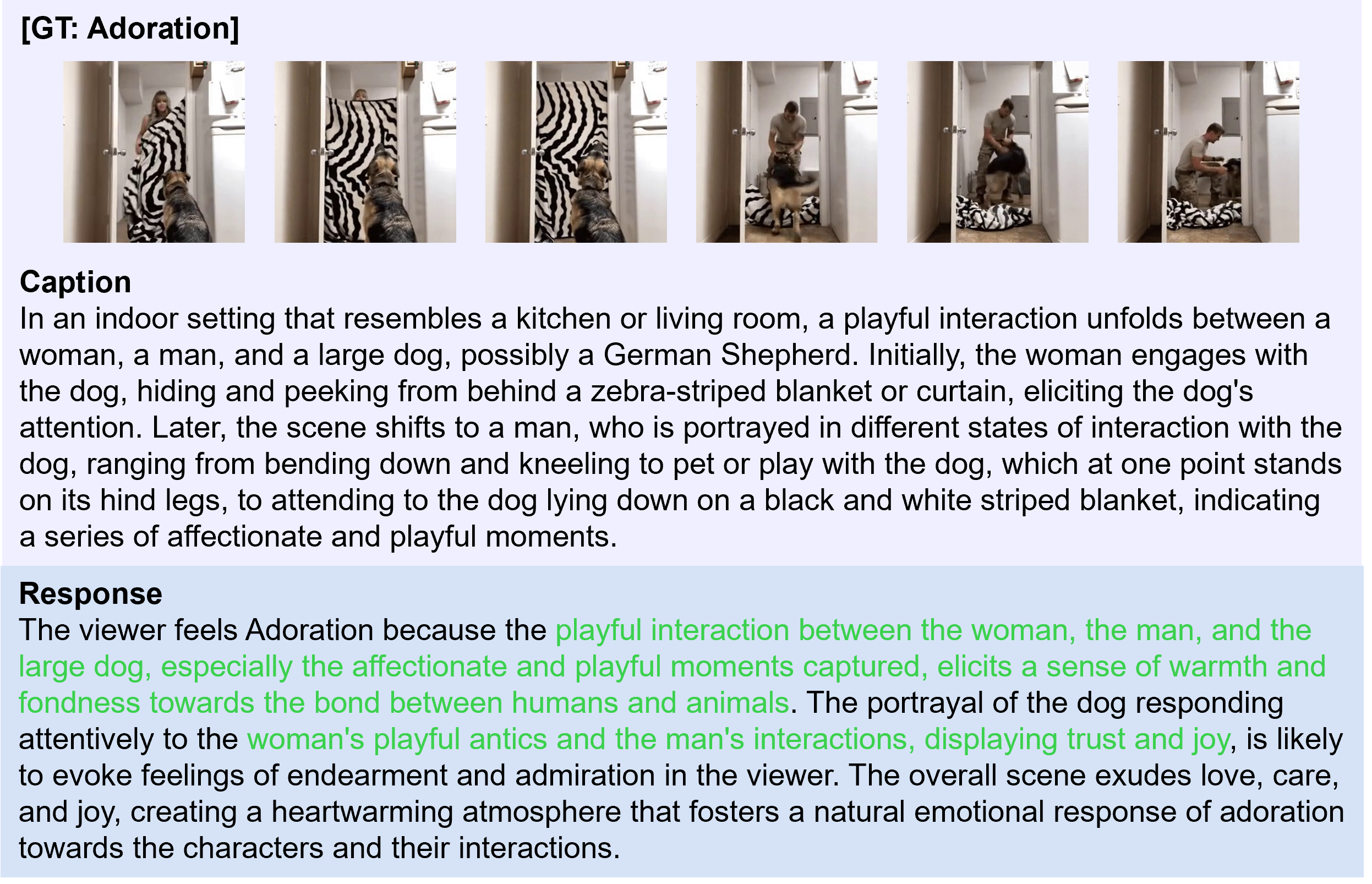} 
        \label{fig:sub4}
    \end{subfigure}
\end{figure}

\begin{figure}[h]
    \centering
    \begin{subfigure}[b]{\textwidth}
        \centering
        \includegraphics[width=\textwidth]{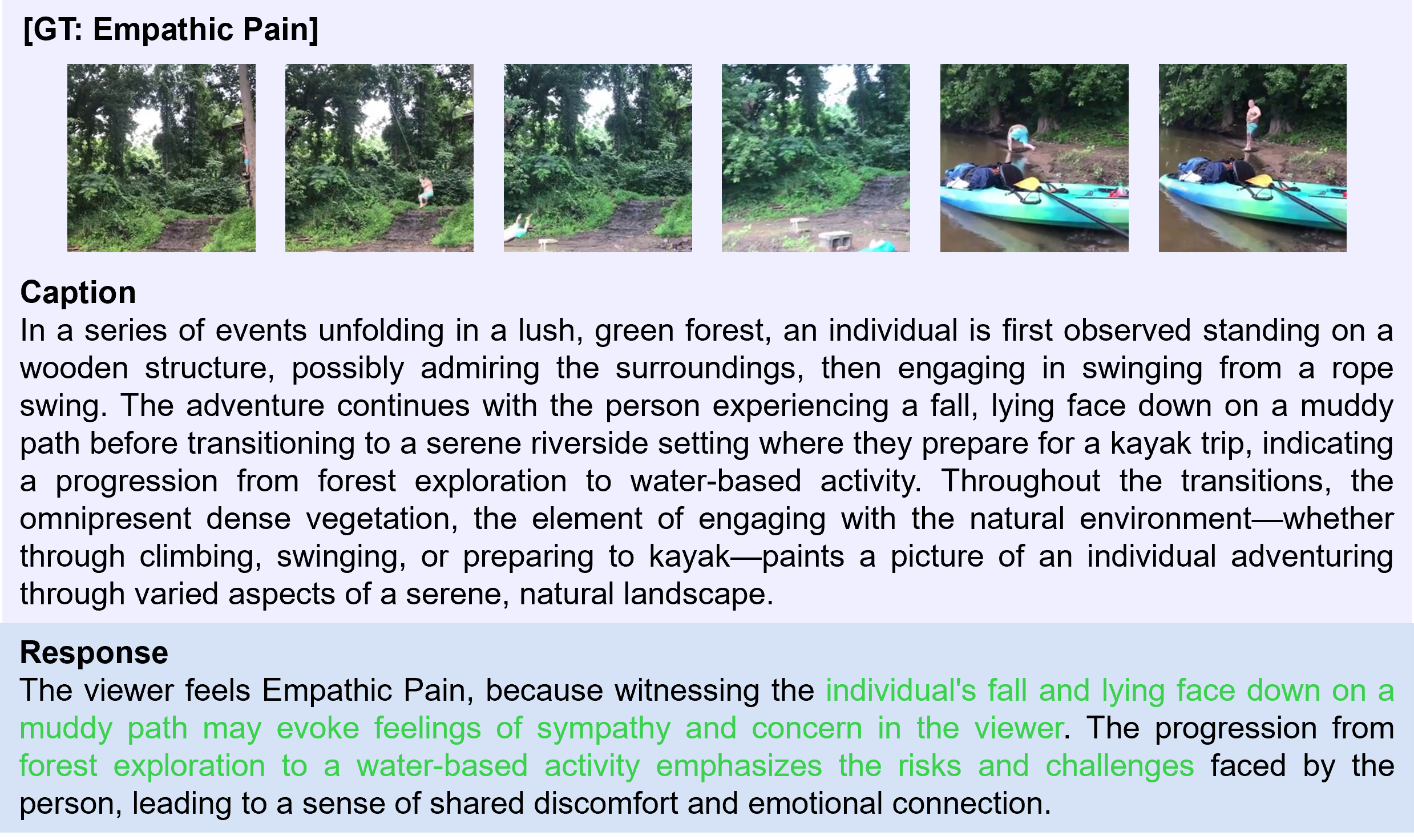} 
        \label{fig:sub5}
    \end{subfigure}
    \hfill
    \begin{subfigure}[b]{\textwidth}
        \centering
        \includegraphics[width=\textwidth]{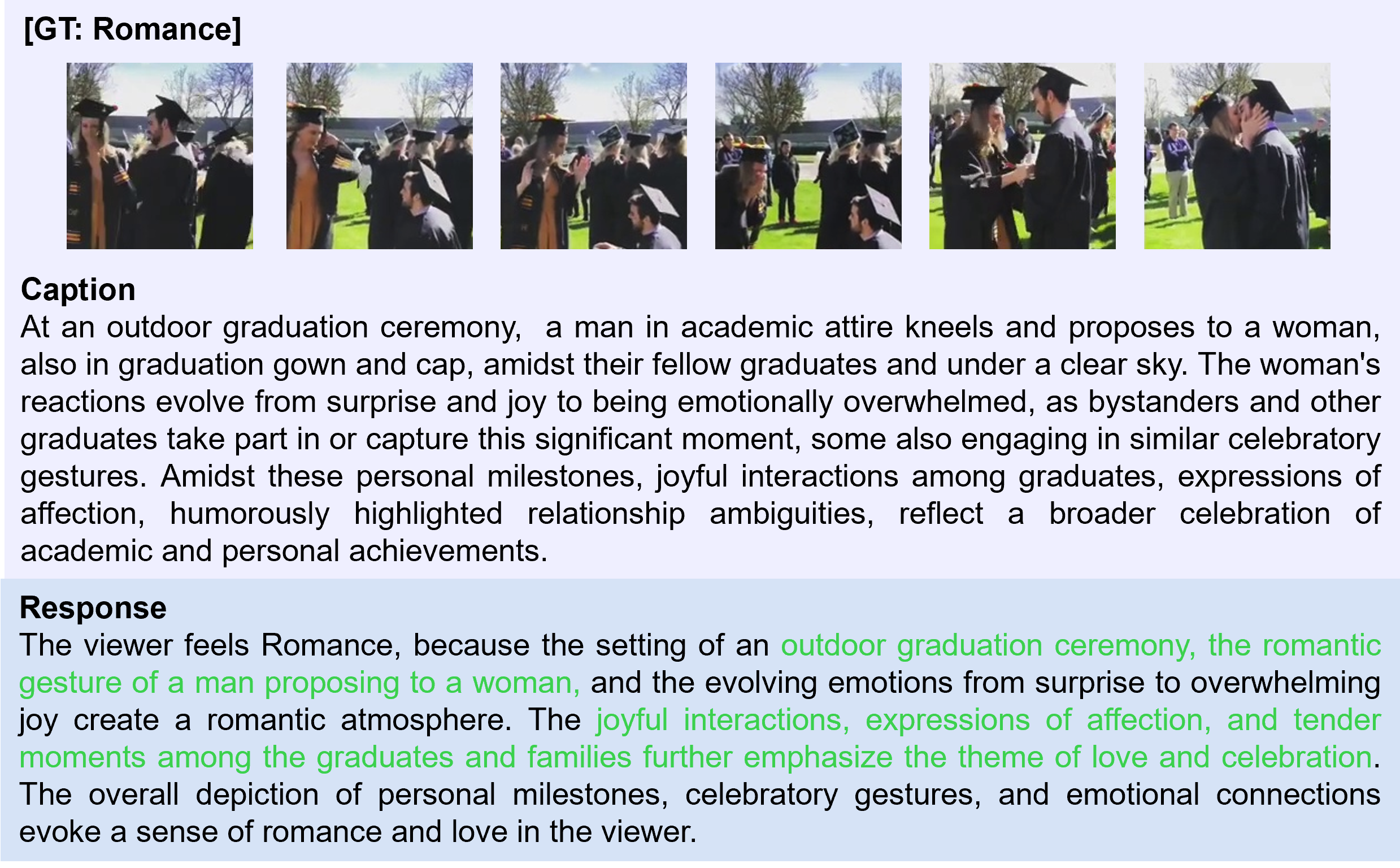} 
        \label{fig:sub6}
    \end{subfigure}
\caption{The examples of VAR data.}
\label{fig:var_examples}
\end{figure}

\end{document}